\pdfoutput=1
\documentclass[11pt]{article}

\usepackage[final]{acl}

\usepackage{times}
\usepackage{latexsym}
\usepackage{graphicx}
\usepackage{arydshln}
\PassOptionsToPackage{table}{xcolor}
\usepackage{amsmath}
\usepackage{relsize}
\usepackage{enumitem}
\usepackage{booktabs}
\usepackage{multirow}
\usepackage{xcolor}
\usepackage{colortbl}
\usepackage{fancyvrb}
\usepackage{tcolorbox}
\usepackage{algorithm}
\usepackage{algorithmic}
\usepackage{color}
\usepackage{hyperref}
\usepackage{url}
\usepackage{multirow}
\usepackage{amsfonts,amssymb}
\usepackage{natbib}
\usepackage{makecell}
\usepackage{tabularx}
\PassOptionsToPackage{table}{xcolor}
\usepackage{booktabs}
\usepackage{listings}
\usepackage{listings}
\usepackage{xcolor}
\usepackage{tabularx}
\usepackage{xcolor}
\usepackage{booktabs}
\usepackage{geometry}
\lstdefinelanguage{json}{
    basicstyle=\ttfamily\footnotesize,
    numbers=none,
    numberstyle=\tiny\color{gray},
    stepnumber=1,
    numbersep=5pt,
    showstringspaces=false,
    breaklines=true,
    frame=single,
    backgroundcolor=\color[gray]{0.95},
    literate=
     *{0}{{{\color{blue}0}}}{1}
      {1}{{{\color{blue}1}}}{1}
      {2}{{{\color{blue}2}}}{1}
      {3}{{{\color{blue}3}}}{1}
      {4}{{{\color{blue}4}}}{1}
      {5}{{{\color{blue}5}}}{1}
      {6}{{{\color{blue}6}}}{1}
      {7}{{{\color{blue}7}}}{1}
      {8}{{{\color{blue}8}}}{1}
      {9}{{{\color{blue}9}}}{1}
      {:}{{{\color{black}:}}}{1}
      {,}{{{\color{black},}}}{1}
      {"}{{{\color{red}"}}}{1},
}
\usepackage[T1]{fontenc}
\usepackage[utf8]{inputenc}
\definecolor{lightblue}{RGB}{215, 235, 250}

\usepackage{microtype}

\usepackage{inconsolata}

\usepackage{graphicx}

\title{CogDual: Enhancing Dual Cognition of LLMs via Reinforcement Learning with Implicit Rule-Based Rewards}

\author{%
  Cheng Liu$^{1,2}$$^{\ast}$, Yifei Lu$^{1,3}$$^{\ast}$, Fanghua Ye$^{1}$$^{\dagger}$, Jian Li$^{1}$$^{\dagger}$, Xingyu Chen$^{1}$ \\
  \textbf{Feiliang Ren$^{3}$, Zhaopeng Tu$^{1}$, Xiaolong Li$^{1}$}\\
  $^1$Hunyuan AI Digital Human, Tencent, Shenzhen, China \\
  $^2$The Chinese University of Hong Kong, Shenzhen, China \\ 
  $^3$Northeastern University, Shenyang, China \\
  \texttt{chengliu2@link.cuhk.edu.cn}, \texttt{lyfei1126@gmail.com} \\
  \texttt{\{fanghuaye, jackjianli\}@tencent.com} \\
}

\begin{document}
\maketitle
\footnotetext[1]{$^{\ast}$ Work done during an internship at Tencent Hunyuan.}
\begingroup
\renewcommand\thefootnote{}\footnotetext{$^{\ast}$ Equal Contribution.}
\renewcommand\thefootnote{}\footnotetext{$^{\dagger}$ Corresponding Author.}
\endgroup

\begin{abstract}
% Role-Playing Language Agents (RPLAs) have emerged as a significant application direction for Large Language Models (LLMs). Existing approaches typically rely on prompt engineering or supervised fine-tuning to enable models to imitate character behaviors in specific scenarios, but they often overlook the underlying cognitive mechanisms driving these behaviors. 
% Inspired by cognitive psychology, we propose \textbf{CogDual}, an RPLA that adopts a \textit{cognize-then-respond} reasoning paradigm. By jointly modeling both external situational awareness and internal self-awareness, CogDual generates behaviors with enhanced psychological consistency and contextual alignment. We further design Inference-Conditioned Likelihood Gain Reward and Latent Semantic Alignment Reward, both suitable for general text-based tasks, and use them within a reinforcement learning framework to optimize model performance. Extensive experiments on the CoSER benchmark, as well as on Cross-MR and LifeChoice, show that CogDual outperforms existing baselines and demonstrates strong generalization across diverse role-playing tasks.

Role-Playing Language Agents (RPLAs) have emerged as a significant application direction for Large Language Models (LLMs). Existing approaches typically rely on prompt engineering or supervised fine-tuning to enable models to imitate character behaviors in specific scenarios, but often neglect the underlying \emph{cognitive} mechanisms driving these behaviors. Inspired by cognitive psychology, we introduce \textbf{CogDual}, a novel RPLA adopting a \textit{cognize-then-respond } reasoning paradigm. By jointly modeling external situational awareness and internal self-awareness, CogDual generates responses with improved character consistency and contextual alignment. To further optimize the performance,
we employ reinforcement learning with two general-purpose reward schemes designed for open-domain text generation. Extensive experiments on the CoSER benchmark, as well as Cross-MR and LifeChoice, demonstrate that CogDual consistently outperforms existing baselines and generalizes effectively across diverse role-playing tasks.
% we employ reinforcement learning with two customized text-based rewards.

% Inference-Conditioned Likelihood Gain Reward and Latent Semantic Alignment Reward—both suitable for general text-based tasks—are integrated within a reinforcement learning framework. 

\end{abstract}

\section{Introduction}
With the rapid advancement of Large Language Models (LLMs), recent years have witnessed a surge of research on role-playing \cite{hdp, rolecraft, role-paly-survey, role_play_survey2}. Role-Playing Language Agents (RPLAs) are designed to equip LLMs with human-like capabilities, enabling them to emulate specific characters across diverse scenarios, while exhibiting behaviors and expressions consistent with the character's profile and context \cite{characterglm}.

\begin{figure}[!ht]
    \centering
    \includegraphics[width=\linewidth]{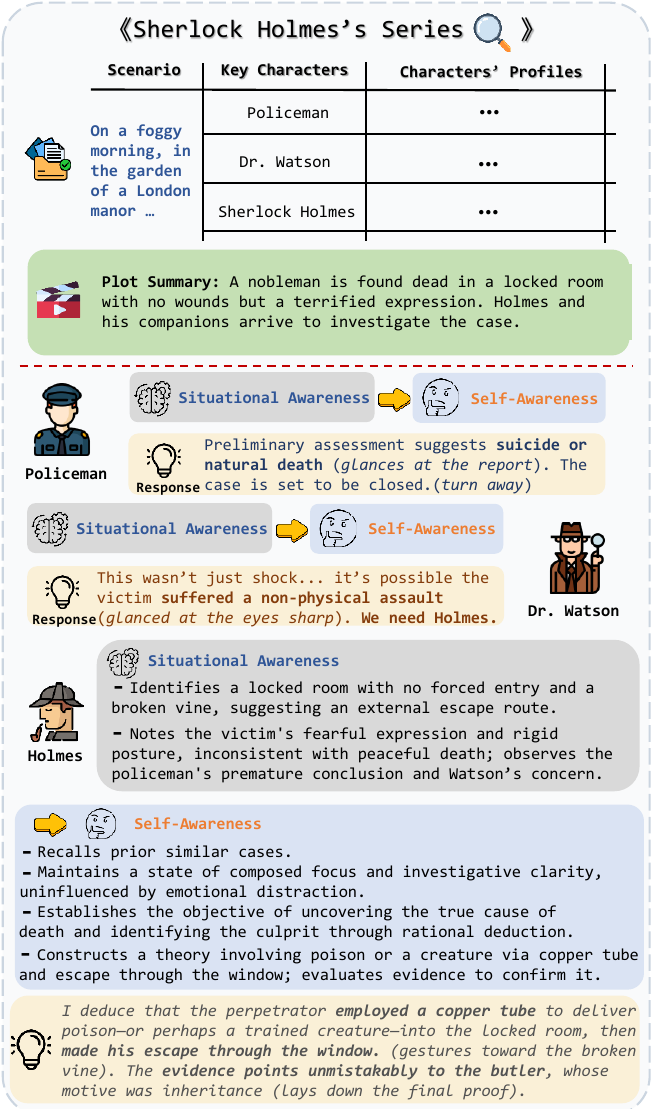}
    \caption{An example of CogDual. Before a character delivers a formal response, it undergoes a \emph{dual cognitive reasoning process}, encompassing external Situational Awareness and internal Self-awareness.}
    \label{fig:example}
    \vspace{-0.3in}
\end{figure}

Previous efforts have primarily focused on constructing role-playing evaluation benchmarks \cite{socialbench, charactereval} and improving model performance through prompt engineering or Supervised Fine-Tuning (SFT). These approaches typically assess whether the model's responses align with a character's profile in specific scenarios. Additionally, some studies employ multiple-choice formats to assess the model to infer motivations \citep{cross_mr}, predict behaviors \citep{LifeChoice}, or analyze psychological states \citep{InCharacter}, thereby quantifying character consistency and fidelity. However, these approaches overlook a critical dimension: \textit{as anthropomorphized agents, RPLAs should engage in cognitive processes involving both situational and self-awareness} rather than merely replicating superficial linguistic patterns or behavioral tendencies.

From the perspective of cognitive psychology \citep{cognition_logic, clark1991grounding, tomasello2010origins}, human role-related behaviors emerge from an integrated cognitive process involving environmental perception, others' behaviors, and introspection of one’s emotions and intentions. This cognitive process plays a crucial role prior to action generation. 
% Building on this foundation, we propose \textbf{CogDual}, a RPLA that incorporates dual cognitive modeling \textcolor{red}{briefly explain what is dual cog, particularly what  "dual" refers to} and embeds the \textit{cognize-then-respond} paradigm into its reasoning process. The goal is to enhance contextual relevance and psychological consistency of responses through cognition-driven generation, thereby enabling improved performance on role-playing tasks.
Building on this foundation, we propose \textbf{CogDual}, a RPLA that incorporates dual cognitive modeling, 
combining outward Situational Awareness and inward Self-Awareness, and embedding a \textit{cognize-then-respond} paradigm into its reasoning process, as illustrated in Figure \ref{fig:example}. By prioritizing cognition-driven generation, CogDual enhances both contextual relevance and psychological consistency in responses, ultimately improving performance on role-playing tasks.

% In addition, motivated by insights and considerations from general-purpose text modeling, we design two broadly applicable reward schemes: 
Motivated by the need to adapt reward modeling for general-purpose text generation, we design two broadly applicable reward schemes: (1) the Inference-Conditioned Likelihood Gain (ICLG) Reward, which quantifies how the intermediate cognitive steps improve response likelihood, and (2) the Latent Semantic Alignment (LSA) Reward, which assesses semantic similarity between generated responses and gold-standard references. Based on these reward designs, we employ reinforcement learning to enhance CogDual’s performance over the supervised fine-tuning baseline. 

In contrast to contemporary studies such as \citet{PCL} and \citet{MIRROR}, which also explore strategies to enhance the reasoning capabilities of RPLAs, our approach distinguishes itself by emphasizing the construction of a comprehensive character cognition process before response generation.  Unlike their fragmented self-questioning or isolated mental state simulation, our dual cognitive reasoning process generates coherent, contextually grounded responses by tightly aligning psychological dynamics with narrative context.

The contributions of this work are as follows:
\begin{itemize}[leftmargin=8pt, topsep=2pt,itemsep=-1pt]

\item We formalize the \textit{cognize–then–respond} paradigm for RPLAs and propose CogDual, the first agent to implement dual cognitive modeling through \textbf{Situational Awareness} and \textbf{Self-Awareness}, providing a more psychologically plausible simulation of human-like behavior generation.

\item We design two reward schemes and demonstrate their effectiveness through reinforcement learning on the CoSER benchmark~\citep{coser}, achieving up to a 9.24\% average improvement over baseline. The proposed reward design may serve as a reference for future research on evaluating text generation in general-domain applications.

\item Through extensive experiments on Cross-MR \citep{cross_mr} and LifeChoice \citep{LifeChoice} benchmarks, we show CogDual's superior cross-task transferability, outperforming all baseline methods.

\end{itemize}

\section{Related Work}
\subsection{Role-Playing Language Agents}

Early investigations into RPLAs centered on character understanding, including character prediction from narrative texts and movie scripts \citep{tell_story, fewshot_character_understanding_movies}. With advances in LLMs, recent studies have extended RPLAs to facilitate character imitation through instruction-based reasoning and supervised fine-tuning, especially in dialogue and knowledge-intensive tasks \citep{character_llm, role_llm, coser}. Beyond imitation, a growing body of work has shifted focus toward evaluating the internal coherence of character-driven behaviors. Studies such as \citep{cross_mr, LifeChoice, InCharacter} have introduced evaluative frameworks incorporating motivation recognition, persona-driven decision making, and psychological evaluation, allowing for a more nuanced analysis of the character consistency and behavioral plausibility of RPLAs.

\subsection{LLM-Based Cognitive Modeling}
Recent studies have increasingly explored the cognitive capacities of LLMs, particularly their ability to exhibit human-like behaviors in dialogic settings~\citep{lamda, generate_agent}. This includes alignment with traits such as self-awareness~\citep{reflexion}, emotion understanding~\citep{empathetic}, intent recognition~\citep{spc}, and deliberative reasoning~\citep{cot, r1}. These abilities are often evaluated in interactive contexts like multi-agent simulations~\citep{camel}, narrative generation~\citep{toward_immersion_drama}, role-playing~\citep{role-paly-survey}, and chatbot systems~\citep{longmemeval, press}. However, recent work highlights that LLMs lack internal psychological states and intrinsic motivations, limiting the depth of their cognitive behaviors~\citep{can_llm_simulation_reflect_human}. Our work adopts a cognitive psychology perspective to more rigorously define and examine LLM cognition in role-play settings.

\subsection{Reasoning Techniques in LLMs}

Recent research has shifted focus from train-time to test-time scaling, with notable success across various tasks such as math problem solving \citep{qwen25_math, s2r}, logical puzzle reasoning \citep{logic_rl} and tool-integrated reasoning \citep{codetool, tool_rl, re_tool}. However, \citet{reasoning_does_not_improve} has highlighted the limitations of reasoning-augmented models \citep{o1, r1} in role-playing scenarios. These models often suffer from stylistic drift between their reasoning traces and character-based generation, thereby undermining the coherence and consistency required for effective role enactment in RPLAs. Our study aims to enhance the generalizability of RPLAs across tasks and domains across various standard benchmarks by reinforcing reasoning process through a cognitively grounded template.

\section{Methodology}

\subsection{Cognition-Driven Reasoning Paradigm}
\begin{quote}
\textit{``Cognition is the activity of knowing: the acquisition, organization, and use of knowledge.''    --- Neisser, 1967}
\vspace{-0.1in}
\end{quote}

This foundational perspective highlights cognition as the driving force behind meaningful communication, rather than a passive background process. While current LLM-based RPLAs can produce fluent utterances, they often overlook the cognitive mechanisms essential to genuine human interaction \citep{cognition_logic, clark1991grounding}.
Motivated by this, we propose a \textbf{cognition-driven reasoning paradigm} for RPLAs, which explicitly embeds cognitive reasoning between perception and response to simulate the psychological steps a human character might take. \citet{tomasello2010origins} shows that individuals interpret environmental and social cues through mental representations, which guide intentional actions, making the transition from external to internal cognition central to human communication.
We thus focus on dual cognition, progressing from external perception to internal reflection. By modeling this cognitive transition, we propose \textbf{CogDual}, which enables RPLAs to generate dual cognition before responding.

\subsection{Preliminaries}
To formally ground the cognition-driven reasoning paradigm introduced above, we first define the key notations and basic concepts used throughout this work. A multi-party dialogue setting is defined over a set of characters, denoted as $\mathcal{O} = \{ o_1, o_2, \dots, o_{|\mathcal{O}|} \}$. Formally, let  $\mathcal{M}$ represent an LLM simulating a specific character $c \in \mathcal{O}$ in a dialogue scene. The model has access to the character’s profile $\mathcal{P}_{c}$, a global scene description $\mathcal{S}$, which may include the current task, storyline, and other elements, and a historical dialogue context $\mathcal{D}_t = \{ d_1, d_2, \dots, d_t \}$, where each $d_i$ represents an utterance, an action, or a thought from a certain character at turn $i$.

The objective of CogDual is to incorporate dual cognition to establish a \textit{cognize-then-respond} pattern. At each time step $t$, $\mathcal{M}$ first performs cognition, forming an internal thinking of the situation, other characters, and itself, and then generates a response. This process is formalized as:
\begin{equation}
c_{t}, d_{t} = \mathcal{M}\left(\mathcal{P}_{c}, \mathcal{S}, \mathcal{O}, \mathcal{D}_{t-1}\right), \quad \mathcal{D}_{0} = \emptyset,
\vspace{-0.05in}
\end{equation}
where $c_t$ denotes the dual cognitive reasoning process at turn $t$, and $d_t$ is the generated response conditioned on $c_t$ and the given inputs. Compared to previous works \citep{role_llm, charactereval, coser} that directly generate $d_t$, our study requires LLMs to perform explicit cognitive thinking before response generation, producing structured representations of the current environment, other characters, and the agent's own state. This mechanism is designed to enhance the model’s contextual understanding in complex scenarios, while improving the coherence and interpretability of character behavior.

\begin{figure*}[!t]
    \centering
    \includegraphics[width=1\linewidth]{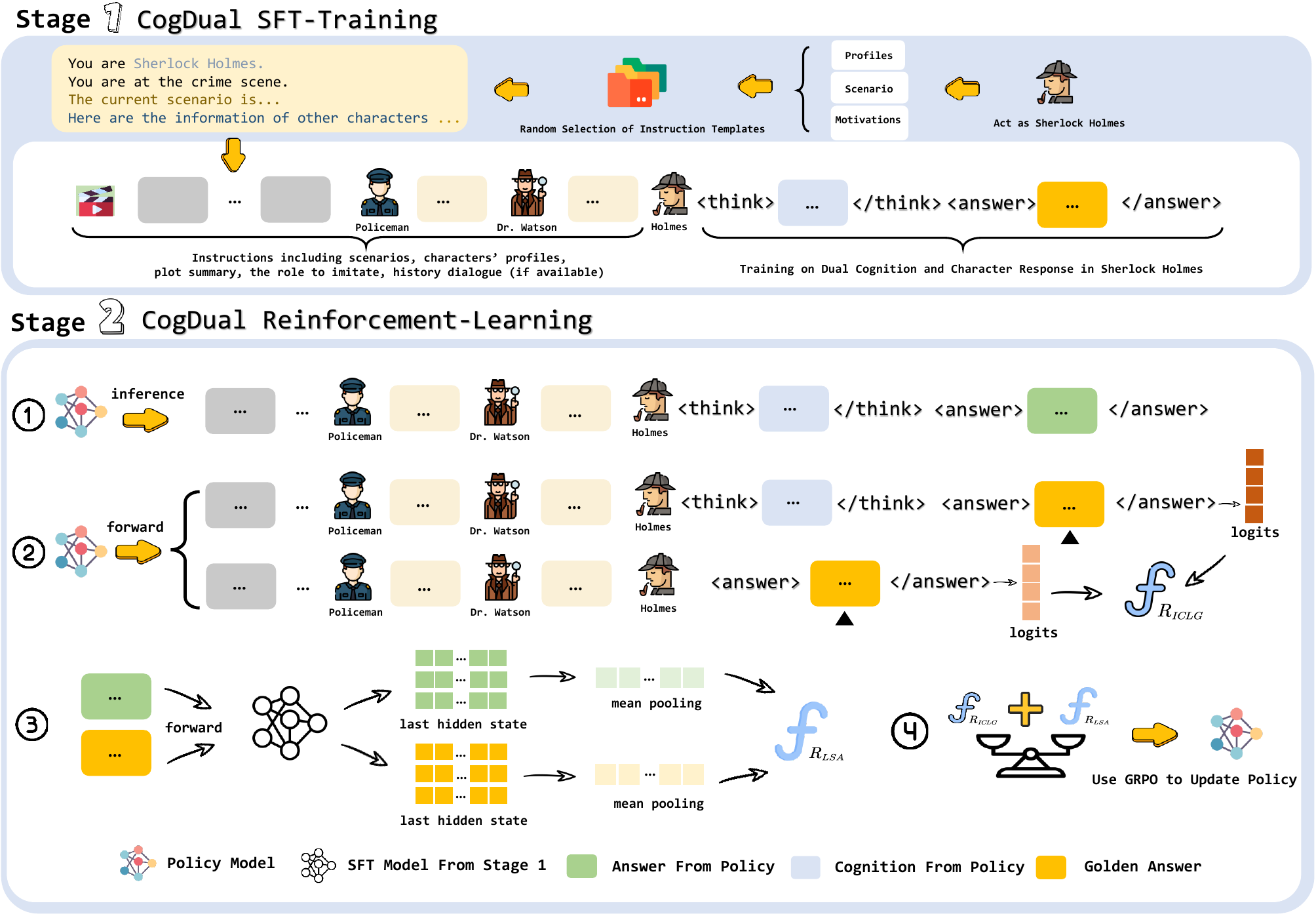}
    \caption{Overview of CogDual training. (1) Stage 1: Supervised fine-tuning using the role-specific dual cognitive reasoning process and corresponding response generated by the RPLA. (2) Stage 2: Reinforcement learning with GRPO, optimized based on the ICLG Reward and the LSA Reward.}
    \label{fig:overview}
    \vspace{-0.2in}
\end{figure*}

\subsection{Dual Cognition of RPLAs}
\label{dual_cognitive_frame}
% \textcolor{green}{need some reference}

In this part, we detail the structure of the Dual Cognition of RPLAs, which consists of two key components: \textbf{Situational Awareness} and \textbf{Self-Awareness}, forming a reasoning process that flows from the outer environment to the inner self. 

\paragraph{Situational Awareness}
Situational Awareness refers to the RPLA's ability to perceive and interpret the environment and other characters within a dialogue scene. It consists of two components:
\textbf{(1) Environmental Perception (EP):} Extracting salient cues from spatial layout, temporal shifts, and scene dynamics, such as changes in atmosphere, character positions or expressions, and dialogue interruptions. These form the initial layer of cognitive input, grounding the agent’s understanding of the unfolding situation.
\textbf{(2) Perception of Others:} Comprising three subcomponents:
\begin{itemize}[leftmargin=8pt, topsep=0pt, itemsep=0pt]
\item \textbf{Behavior Analysis (BA):} Identifying key actions or speech patterns from others that may affect the agent’s response;
\item \textbf{Emotion Analysis (EA):} Inferring emotional states from behavior and context, and assessing their impact on the agent;
\item \textbf{Intention Analysis (IA):} Inferring others’ intentions to guide the agent’s reactions.

\end{itemize}

This process can be formally represented as: 
\begin{equation}
% \vspace{-0.05in}
SA = \langle \text{EP}, \text{BA}, \text{EA}, \text{IA} \rangle,
\vspace{-0.1in}
\end{equation}
where $SA$ denotes \textbf{Situational Awareness} formed through hierarchical perception and interpretation.

\paragraph{Self-Awareness} 
Self-Awareness forms a core component of the cognitive architecture in RPLAs, enabling introspection and adaptive decision making. It comprises four interrelated elements:
\begin{itemize}[leftmargin=8pt, topsep=0pt,itemsep=0pt]

\item \textbf{Key Memory Activation (KMA):} Recalling autobiographical or episodic memories relevant to the current context;

\item \textbf{Self-Emotion (SE):} Recognizing and evaluating internal emotional states that influence perception and behavior;

\item \textbf{Self-Intention (SI):} Maintaining context-driven goals that guide actions;

\item \textbf{Internal Strategy (IS):} Integrating memory, emotion, and intention into coherent reasoning for planning and outcome anticipation.
\end{itemize}

This process can be formally represented as:
\begin{equation}
SA_{self} = \langle \text{KMA}, \text{SE}, \text{SI}, \text{IS} \rangle,
    \vspace{-0.1in}
\end{equation}
where $SA_{self}$ denotes the \textbf{Self-Awareness} formed through the agent's self-cognition.

\subsection{Dual Cognition Behavior Learning}

We propose two approaches for dual cognition: a cognitive-based Chain-of-Thought (CB-CoT) prompting method and a two-stage training framework, as shown in Figure~\ref{fig:overview}. This section focuses on the latter; CB-CoT is detailed in Appendix~\ref{appendix:cb-cot}. For supervised training, we construct a dataset $\mathcal{D}_{\text{SFT}}$ with dual cognition trajectories (see Appendix~\ref{appendix:details_on_CoSER}).

\subsubsection{Stage 1: Supervised Fine-tuning for CogDual Initialization}
Once the dual cognition training dataset $\mathcal{D}_{SFT}$ is ready, we initialize cognitive behavior modeling of the LLM via SFT, optimizing the following negative log-likelihood objective:
\begin{equation}
    \mathcal{L}_{\text{SFT}} = - \mathbb{E}_{(x, y) \sim \mathcal{D}_{SFT}} \sum_{i=1}^N \log \pi(y_i \mid x, y_{<i}),
\end{equation}
where $\pi$ denotes the policy of $\mathcal{M}$, $i$ is the token index, and $x = \{\mathcal{P}_{c}, \mathcal{S}, \mathcal{O}, \mathcal{D}\}$ represents the full input context composed of the character’s profile, a global scene description, a set of characters, and dialogue history, respectively. 

\begin{table*}[!t]
    \centering
    % \small
    \resizebox{\textwidth}{!}{
    \begin{tabular}{lcccccc}
    \toprule
     \textbf{Models}  & \textbf{Methods} & \textbf{Storyline Consistency}  & \textbf{Anthropomorphism}  & \textbf{Character Fidelity} & \textbf{Storyline Quality}& \textbf{Average} \\
     \midrule
     \rowcolor[gray]{0.9} % 设置灰色背景，0.9 表示浅灰色
     \multicolumn{7}{c}{\textbf{Closed-Source LLMs}} \\
    \multirow{3}{*}{GPT-3.5-Turbo-0613} 
        & Vanilla & 53.37 & \underline{39.53} & \underline{35.99} & 70.28 & 49.79 \\
        & + CoT & \underline{55.75} & 39.21 & 35.36 & \textbf{72.26} & \underline{50.64}  \\ 
        & + CB-CoT & \textbf{59.84} & \textbf{46.23} & \textbf{44.50} & \underline{70.71} & \textbf{55.32} \\
        \midrule
    \multirow{3}{*}{GPT-4o} 
        & Vanilla & \underline{58.93} & 43.14 & \textbf{41.62} & \underline{75.36} & \underline{54.76} \\
        & + CoT & 58.65 & \textbf{44.37} & 38.18 & \textbf{77.72} & 54.73 \\
        & + CB-CoT & \textbf{59.80} & \underline{44.12} & \underline{40.71} & 74.78 & \textbf{54.85} \\
        \midrule
     GPT-o1-Preview
        & Vanilla & 59.47 & 46.81 & 40.54 & 77.80 & 56.16 \\
     \midrule
     \rowcolor[gray]{0.9}
    \multicolumn{7}{c}{\textbf{Open-Source LLMs}} \\
    \multirow{3}{*}{LLaMA3.1-70B-Instruct} 
        & Vanilla & 54.63 & 45.54 & 37.99 & 72.62 & 52.69 \\
        & + CoT & 55.36 & 46.96 & 35.80 & 72.92 & 52.76 \\
        & + CB-CoT & \underline{57.74} & \underline{49.13} & 38.57 & \underline{74.89} & 55.08 \\
        & + CoSER & 56.58 & \textbf{49.27} & \underline{41.46} & \textbf{75.84} & \underline{55.79} \\
        & + LongCoT & \textbf{64.18} & 41.42 & \underline{44.01} & 72.96 & 55.64 \\
        \rowcolor{lightblue!100} & + \textbf{CogDual-SFT(ours)} & 57.60 & 48.02 & \textbf{48.55} & 72.75 & \textbf{56.73} \\
    \midrule
    \multirow{7}{*}{Qwen2.5-7B-Instruct} 
        & Vanilla & \underline{59.86} & 42.03 & 41.45 & 62.32 & 51.41 \\
        & + CoT & 55.76 & 37.21 & 36.5 & 61.80 & 47.82 \\
        & + CB-CoT & 56.88 & 44.91 & 39.11 & 62.46 & 50.84 \\
        & + CoSER & 56.44 & 44.27 & 41.79 & 68.95 & 52.86 \\
        & + LongCoT & 58.83 & 40.56 & \underline{45.05} & 61.52 & 51.48 \\
        \rowcolor{lightblue!100} & + \textbf{CogDual-SFT(ours)} & 58.36 & \textbf{46.95} & 44.99 & \underline{71.72} & \underline{55.51} \\
        \rowcolor{lightblue!100} & + \textbf{CogDual-RL(ours)} & \textbf{59.94} & \underline{46.64} & \textbf{46.95} & \textbf{73.97} & \textbf{56.88} \\
    \midrule
        \multirow{5}{*}{LLaMA3.1-8B-Instruct} 
        & Vanilla & 48.17 & 36.58 & 26.98 & 63.70 & 43.85 \\
        & + CoT & 50.14 & 40.39 & 27.95 & 64.27 & 45.69 \\
        & + CB-CoT & 52.79 & 41.44 & 27.72 & 65.03 & 46.74 \\
        & + CoSER & 52.78 & 43.96 & 37.47 & 70.60 & 51.20 \\
        & + LongCoT & \underline{59.49} & 40.85 & \underline{44.98} & 63.47 & 52.20 \\
        \rowcolor{lightblue!100} & + \textbf{CogDual-SFT(ours)} & 55.99 & \textbf{46.92} & 43.78 & \textbf{75.07} & \underline{55.44} \\
        \rowcolor{lightblue!100} & + \textbf{CogDual-RL(ours)} & \textbf{60.10} & \underline{45.89} & \textbf{48.82} & \underline{73.08} & \textbf{56.97} \\
    \bottomrule
    \end{tabular}}
    \caption{
    The performance of CogDual and baselines on the most comprehensive role-playing benchmark, CoSER.
    \textbf{Vanilla} refers to models without any method.
    \textbf{CB-CoT} denotes our proposed cognitive-based Chain-of-Thought prompting method (see Appendix~\ref{appendix:cb_cot} for details).
    \textbf{CogDual-SFT} is the fine-tuned model from stage 1, while \textbf{CogDual-RL} is trained with our proposed RL.
    The best results are in \textbf{bold}, suboptimal ones are \underline{underlined}.
    }
    \label{tab:main}
    % \vspace{-0.2in}
\end{table*}
\subsubsection{Stage 2: Reinforcement Learning with Two Implicit Rule-Based Rewards}

To further improve generalization after cognitive behavior initialization, we introduce a reinforcement learning (RL) stage with two implicit rule-based reward mechanisms. Unlike conventional open-domain RLHF pipelines that rely on an external reward model, which requires large-scale data collection and costly human annotation or on explicit rule-based checks typically applicable only to code or math tasks, our reward signals are derived entirely from the model’s internal outputs and a frozen reference policy. This design incurs no additional annotation cost and remains broadly applicable to any RPLA setting. Existing general-purpose reward models (e.g., CharacterEval~\citealp{charactereval}) address only limited aspects of the task, offer constrained interpretability, and have not been validated in RL settings. 

Given these limitations, we propose two complementary implicit reward mechanisms and optimize the policy using Grouped Reward Policy Optimization (GRPO;~\citealp{deepseek-math}).

\paragraph{ICLG: Rewarding Reasoning Utility via Likelihood Gain}
\label{ICLG}
Inspired by LATRO~\citep{latro}, which uses the $\log \pi_\theta(y \mid x \oplus z)$ of a reasoning-augmented output as a reward, where $z$ denotes an intermediate rationale. We introduce Inference-Conditioned Likelihood Gain (ICLG) to promote causal consistency in cognitive reasoning. ICLG directly measures how much explicit reasoning increases the likelihood of producing the correct response, thereby rewarding reasoning traces that effectively support accurate and coherent generation.
Concretely, given a pair $(x, d_{\text{golden}})$ consisting of a prompt $x$ and its reference response $d_{\text{golden}}$, the policy model performs a dual cognition rollout on input $x$, producing a reasoning trace $c$ followed by a response $\hat{d}$, i.e., a trajectory $(c, \hat{d})$. The ICLG reward evaluates, on a per-token basis, how conditioning on the model’s own cognition $c$ improves the likelihood of generating the $d_{\text{golden}}$.

{\small
\begin{align}
R_{\text{ICLG}}(x, d_{\text{golden}}, c) &=
\left( \frac{\pi_\theta(d_{\text{golden}} \mid x \oplus c)}{\pi_\theta(d_{\text{golden}} \mid x)} \right)^{1/|d_{\text{golden}}|} \notag \\
&\hspace{-2.5em}= \left( \frac{\prod_{t=1}^{|d_{\text{golden}}|} \pi_\theta(d_t \mid d_{<t}, x \oplus c)}{\prod_{t=1}^{|d_{\text{golden}}|} \pi_\theta(d_t \mid d_{<t}, x)} \right)^{1/|d_{\text{golden}}|}, 
\end{align}
}where $|d_{\text{golden}}|$ denotes the number of tokens in $d_{golden}$. Intuitively, the ICLG encourages reasoning traces that improve fluency and causal coherence while supporting more confident generation.

\paragraph{LSA: Rewarding Semantic Fidelity in Generation}
\label{LSA}
ICLG strongly promotes fidelity but can lead the model to imitate the surface form of
$d_{\text{golden}}$, reducing expressive diversity that is crucial in role-play. 
Hence, we introduce Latent Semantic Alignment (LSA), which rewards outputs
that are semantically—rather than lexically—close to the reference. Unlike conventional token-level objectives~\citep{tokenmatch}, LSA measures the semantic similarity between  $\hat{d}$ and $d_{\text{golden}}$ in the latent space of a frozen reference model, $\pi_{\text{ref}}$ (i.e., the RPLA after SFT). Formally,

\begin{equation}
\small
\begin{aligned}
R_{\text{LSA}}(x, d_{\text{golden}}, \hat{d}) = \cos\left( f_{\text{ref}}(x, d_{\text{golden}}), f_{\text{ref}}(x, \hat{d}) \right),
\end{aligned}
\end{equation}where $f_{\text{ref}}(x, d) = \frac{1}{T} \sum_{t=1}^{T} h_t$ is the mean-pooled representation of the last hidden states ${h_1, \dots, h_T}$, with $T$ as the length of $d$. $\cos(\cdot,\cdot)$ denotes cosine similarity. This removes the need for a separate encoder and uses the semantic space adapted for role-play via SFT. Prior work~\citep{llm-mean-embedding} shows that mean-pooled representations are effective for semantic similarity. Importantly, LSA is more flexible than SFT: it rewards outputs semantically close to the reference, regardless of wording, enabling the model in RL to remain faithful while allowing more natural, diverse expressions.

\paragraph{RL via GRPO with Two Implicit Rule-Based Rewards}
\label{GRPO_Step}
We optimize our policy model using the GRPO algorithm, which is well-suited for non-smooth, high-variance reward scenarios \citep{grpo_suit1, grpo_suit2} as commonly found in reasoning and generation tasks. In our case, we combine the ICLG and LSA rewards via fixed weights $\lambda_{\text{ICLG}}$ and $\lambda_{\text{LSA}}$, $R$ is computed as follows:
\begin{equation}
\begin{aligned}
R(x, d_{\text{golden}}, c, \hat{d}) &= \lambda_{\text{ICLG}} \cdot R_{\mathrm{ICLG}}(x, d_{\text{golden}}, c) \\
& + \lambda_{\text{LSA}} \cdot R_{\mathrm{LSA}}(x, d_{\text{golden}}, \hat{d}).
\end{aligned}
\end{equation}

For each trajectory $(x, d_{\text{golden}}, c, \hat{d})$, we compute the estimated advantage as follows:
\begin{equation}
\small
\begin{aligned}
A(x, d_{\text{golden}}, c, \hat{d}) =
\frac{
R(x, d_{\text{golden}}, c, \hat{d}) - \frac{1}{|\mathcal{B}|} \sum_{j \in \mathcal{B}} R^{(j)}
}{
\sqrt{
\frac{1}{|\mathcal{B}|} \sum_{j \in \mathcal{B}} \left( R^{(j)} - \frac{1}{|\mathcal{B}|} \sum_{k \in \mathcal{B}} R^{(k)} \right)^2
}
},
\end{aligned}
\end{equation}
where $\mathcal{B}$ is the set of trajectories in the current minibatch.
Putting it all together, we minimize the following surrogate loss to update the policy parameters $\theta$ using trajectories collected from the current policy:
\begin{equation}
\small
\begin{aligned}
\mathcal{L}_{\text{GRPO}}(\theta) = -\mathbb{E}_{\substack{
x \sim \mathcal{D_{\text{RL}}} \\
y \sim \pi_{\theta_{\text{old}}}(\cdot | x)
}} \Bigg[
    \frac{1}{G} \sum_{i=1}^G \frac{1}{T_i} \sum_{t=1}^{T_i} \Big\{
        & \min \Big[
            r_{i,t} \hat{A}_{i,t}, \\
        & \hspace{-15.0em} \mathrm{clip}\big(r_{i,t}, 1-\epsilon, 1+\epsilon\big)\, \hat{A}_{i,t}
        \Big]
        -\; \beta D_{\mathrm{KL}}\big[\pi_\theta \| \pi_{\mathrm{ref}}\big]
    \Big\}
\Bigg],
\end{aligned}
\end{equation}
where $T_i$ is the length of the $i$-th generated sequence, $r_{i,t} = \frac{\pi_\theta(\hat{y}_{i,t} \mid x_i, \hat{y}_{i,<t})}{\pi_{\theta_{\text{old}}}(\hat{y}_{i,t} \mid x_i, \hat{y}_{i,<t})}$ is the importance ratio, $\beta$ controls the strength of the KL penalty. \(\mathcal{D}_{\text{RL}}\) denotes the set of prompts used during the RL stage to generate training trajectories.

\section{Experimental Setup}
To evaluate the effectiveness of CogDual, we conduct comprehensive experiments on CoSER \cite{coser} as the main benchmark, and further assess generalization on Cross-MR~\citep{cross_mr} and LifeChoice~\citep{LifeChoice}. 

\subsection{Base Models}
To evaluate the generality of our method across different LLMs, we conduct main experiments on three open-source models: LLaMA3.1-8B-Instruct, Qwen2.5-7B-Instruct and LLaMA3.1-70B-Instruct.
In addition, we apply the prompting method to three proprietary LLMs: GPT-3.5-Turbo, GPT-4o, and o1-preview, representing models specialized for language understanding, multimodal, and advanced reasoning capabilities.

\subsection{Baselines}
\label{baselines}
To evaluate the effectiveness of our approach, we compare against the following strong baselines widely used in role-playing scenarios:

\begin{itemize}[leftmargin=8pt, topsep=2pt,itemsep=0pt]

\item \textbf{Chain-of-Thought (CoT):} We construct a CoT prompting baseline (as shown in Table~\ref{table:cot_system_prompt}) for direct comparison with our cognition-based CoT approach described in Appendix~\ref{appendix:cb_cot}.

\item \textbf{Vanilla SFT with Different Data Constructions:} We compare LLMs fine-tuned on several data configurations: (1) CoSER: the complete CoSER dataset; (2) LongCoT, long-form CoT-style reasoning data constructed from the same source as CogDual (details in Appendix~\ref{appendix:CoT Data Construction}). For fair comparison, the size of LongCoT data is same as the initialization data of CogDual.
% More details on baseline data construction are provided in \textcolor{green}{Appendix xxx}.

\end{itemize}

\subsection{Evaluation Metrics}

Following CoSER, we evaluate simulated conversations using GPT-4o as a critic across four key dimensions:
\textbf{(1) Storyline Consistency:} Assesses alignment between simulated dialogue $\tilde{\mathcal{D}}$ and original $\mathcal{D}$, focusing on whether RPLA responses (emotions, attitudes, behaviors) remain faithful to the narrative context.
\textbf{(2) Anthropomorphism:} Evaluates whether RPLA exhibits human-like behavior in self-identity, emotional depth, persona consistency, and social interaction.
\textbf{(3) Character Fidelity:} Measures how well the RPLA reflects its character, including style, knowledge, personality, behavior, and relationships.
\textbf{(4) Storyline Quality:} Judges overall coherence and fluency, with emphasis on logical flow and narrative development.

% \paragraph{Training Settings}
% The detailed training hyperparameters are reported in \textcolor{green}{Appendixxxx.}
\begin{table}[!t]
    \centering
    % \small
    \resizebox{1\columnwidth}{!}{
    \begin{tabular}{lccc}
    \toprule
     % \textbf{Models} & \textbf{Cross-MR}\citep{cross_mr} &  \textbf{LifeChoice}\citep{LifeChoice} \\
     \textbf{Models} & \textbf{Methods} & \textbf{Cross-MR} &  \textbf{LifeChoice} \\
     \midrule
     \rowcolor[gray]{0.9} % 设置灰色背景，0.9 表示浅灰色
     \multicolumn{4}{c}{\textbf{Closed-Source LLMs}} \\
     \midrule
    GPT-4o & Vanilla & 36.04 & 73.92 \\
    o1-Preview & Vanilla & \textbf{62.98} &  \textbf{80.08} \\
    \midrule
    \rowcolor[gray]{0.9} % 设置灰色背景，0.9 表示浅灰色
    \multicolumn{4}{c}{\textbf{Open-Source LLMs}} \\
    \multirow{5}{*}{LlaMA3.1-8B-Instruct} 
    & Vanilla & 30.15 & 61.10 \\ 
    & + CoSER & 43.39 & 69.54 \\ 
    & + LongCoT & 37.75 & 69.54 \\ 
    & \textbf{+ CogDual-SFT(ours)} & 49.21 & 73.38 \\ 
    & \textbf{+ CogDual-RL(ours)} & 52.81 & 74.15 \\ 
    \midrule
    \multirow{5}{*}{Qwen2.5-7B-Instruct}
    & Vanilla & 54.16 & 68.58 \\ 
    & + CoSER & 56.74 & 67.08 \\ 
    & + LongCoT & 57.19 & 65.43 \\ 
    & \textbf{+ CogDual-SFT(ours)} & 59.66 & 72.63 \\ 
    & \textbf{+ CogDual-RL(ours)} & \underline{60.79} & \underline{74.60} \\ 
    \bottomrule
    \end{tabular}}
    \caption{Accuracy comparison on Cross-MR and LifeChoice. Best results are in \textbf{bold}, while suboptimal ones are \underline{underlined}.}
    \label{tab:genelization}
    \vspace{-0.2in}
\end{table}
\section{Experimental Results and Analyses}
\subsection{Main Results}
Table~\ref{tab:main} shows an overall comparison between CogDual and strong baselines. The results show that:
\begin{itemize}[leftmargin=8pt, topsep=2pt,itemsep=2pt]
\item CogDual consistently improves role-playing performance across all base models. Notably, even without training, our prompting method (CB-CoT) yields substantial gains. After two-stage training, CogDual achieves an 11.65\% boost in \textit{Storyline Quality} for Qwen2.5-7B-Instruct and a 21.84\% gain in \textit{Character Fidelity} for Llama3.1-8B-Instruct, with an average increase of 13.12\%.

\item Generally, CogDual outperforms baselines on most metrics. Notably, Qwen2.5-7B-Instruct with CogDual-RL matches or surpasses o1-preview and even outperforms the much larger Llama3.1-70B-Instruct-CoSER, despite using less than 10\% of the data and only 10,000 RL instances with implicit reward supervision. This highlights CogDual's data and training efficiency.

\item CogDual also clearly outperforms Long-CoT baselines distilled from GPT-4o, even with the same size of SFT data. This demonstrates the effectiveness of CogDual for smaller models in challenging role-play tasks and offers a practical solution for test-time scaling. It also addresses concerns that reasoning-optimized LLMs may be less suitable for role-playing~\cite{reasoning_does_not_improve}.

\end{itemize}

\begin{table*}[!t]
\centering
\large
\resizebox{\textwidth}{!}{
\begin{tabular}{lcc|ccccc|cc}
\toprule
\textbf{Model} & \textbf{$\lambda_{ICLG}$} & \textbf{$\lambda_{LSA}$} & \textbf{Storyline Consistency} & \textbf{Anthropomorphism} & \textbf{Character Fidelity} & \textbf{Storyline Quality} & \textbf{Average} & \textbf{Cross-MR} & \textbf{LifeChoice} \\
\midrule
% \rowcolor[gray]{0.97}
CogDual-SFT & - & - & 55.99 & 46.92 & 43.78 & 75.07 & 55.44 & 49.21 & 73.38 \\
\midrule
\multirow{5}{*}{CogDual-RL}
  & 1.0 & 0.0 & \underline{59.10} & \underline{47.37} & \underline{47.14} & \underline{72.42} & \underline{56.51} & \underline{55.51} & 75.13 \\
  & 0.7 & 0.3 & \textbf{60.10} & {45.89} & \textbf{48.82} & \textbf{73.08} & \textbf{56.97} & \textbf{55.73} & \textbf{78.77} \\
  & 0.5 & 0.5 & 56.31 & 45.20 & 41.54 & {71.04} & 53.52 & 54.38 & \underline{76.17} \\
  & 0.3 & 0.7 & 57.55 & 46.64 & 42.79 & 70.45 & 54.36 & 52.58 & 75.38 \\
  & 0.0 & 1.0 & 57.47 & \textbf{47.63} & 43.24 & 69.38 & 54.43 & 53.71 & 74.41 \\

\bottomrule
\end{tabular}
}
\caption{
Ablation on reward weight combinations. Each RL variant is annotated with its ICLG and LSA weights from Section~\ref{GRPO_Step}. CogDual-SFT and CogDual-RL denote models trained on \texttt{LLaMA3.1-8B-Instruct}. The best results are highlighted in \textbf{bold}, while suboptimal ones are marked with \underline{underline}.
}
\label{tab:reward-ablation}
 % \vspace{-0.2in}
\end{table*}

\paragraph{Implicit Rule-Based Reward RL Analysis.}
We further analyze the effectiveness of the proposed implicit rule-based rewards. As shown in Table~\ref{tab:main}, RL models consistently outperform SFT-only models in both \textit{Storyline Consistency} and \textit{Character Fidelity}, with average improvements of 2.85 and 3.50 points, respectively. 
This indicates that the ICLG reward effectively guides the model to produce reasoning traces that advance the narrative in a causal, coherent manner, while the LSA reward promotes closer alignment between generated actions and the character's intended persona. Notably, Qwen2.5-7B-Instruct with our RL framework achieves the highest overall performance, even surpassing o1-preview on multiple metrics. 
These results demonstrate that our implicit rule-based reward strategy is an efficient and effective alternative to conventional reward modeling for role-play LLMs.

\subsection{Generalization to Other Benchmarks}
We posit that CogDual, through dual cognitive reasoning, demonstrates strong generalization potential and can be extended to other role-playing evaluation benchmarks.
To validate this, we conduct experiments on two well-recognized benchmarks: Cross-MR \citep{cross_mr} and LifeChoice \citep{LifeChoice}. Specifically, Cross-MR requires inferring the motivation behind a character’s decision, while LifeChoice evaluates whether the model can reproduce a character’s original choice based on profile, context, and decision point. Both benchmarks adopt a multiple-choice format, allowing evaluation via \textit{Accuracy}, consistent with their original settings.
To align CogDual with this format, we use GPT-4o to choose the option that is most semantically similar to the response part generated by CogDual(details in Appendix~\ref{appendix:semantic_matching}).
As shown in Table \ref{tab:genelization}, CogDual-equipped LLMs consistently outperform all baselines on both benchmarks. Their performance is also comparable to the strong reasoning model o1-Preview, demonstrating CogDual’s robust generalization.
Notably, the reinforcement learning strategy based on our proposed ICLG and LSA rewards consistently outperforms CogDual-SFT, further validating the effectiveness of our reward design and pushing the upper bound of the model's performance.

\begin{figure}[h]
\centering
\includegraphics[width=1.0\linewidth]{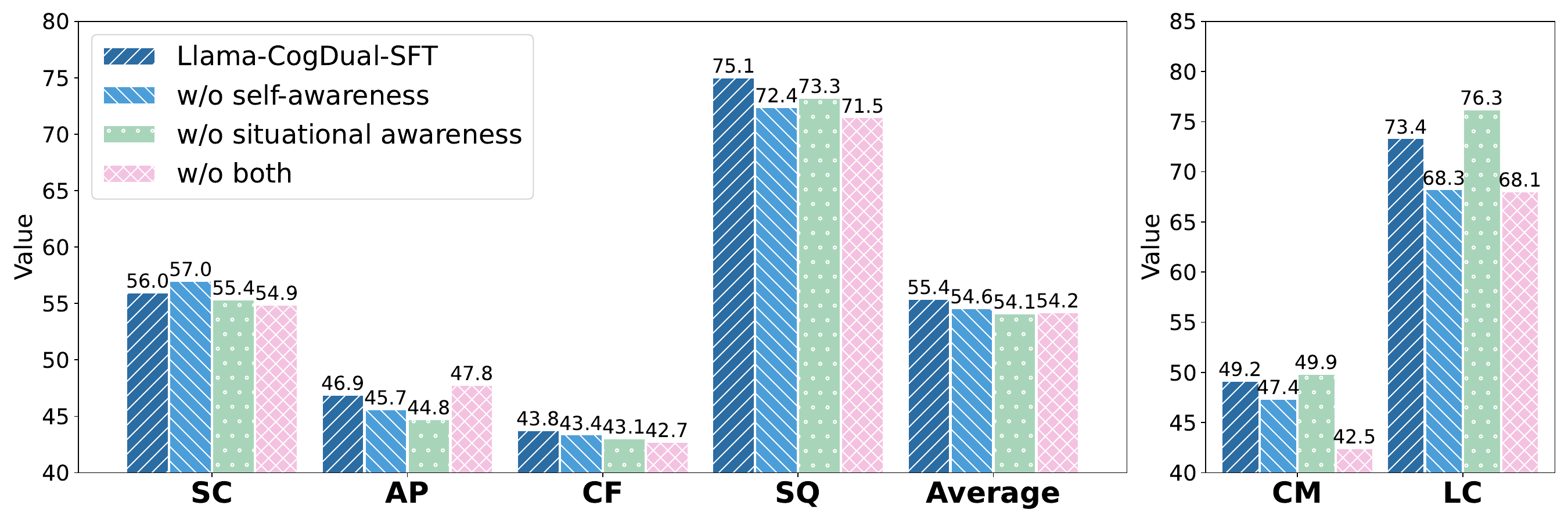}\\
% \vspace{-0.3cm}
    \caption{Ablation study on cognitive components. 
"SC", "AP", "CF", and "SQ" correspond to the four metrics: Storyline Consistency, Anthropomorphism, Character Fidelity, and Storyline Quality, respectively. 
"CM" denotes Cross-MR, and "LC" denotes LifeChoice.}
    \label{fig:cog_ablation}
    % \vspace{cm}
\end{figure}

\subsection{Ablation Study I: Effect of Dual Cognition Components}
We first conduct an ablation study focusing on the effect of dual cognition structures during SFT stage. Figure \ref{fig:cog_ablation} compares four supervision settings: the complete dual cognition model, the removal of self-awareness, the removal of situational awareness, and the removal of both. We find two key results: 

\begin{itemize}[leftmargin=8pt, topsep=2pt,itemsep=0pt]
\item The full dual cognition model provides the most balanced and robust performance, yielding the highest or near-highest scores across primary role-play metrics, including \textit{Storyline Consistency}, \textit{Character Fidelity}, and overall average performance. This result confirms that narrative coherence and stable character portrayal are optimally supported when the model simultaneously reasons over external contexts and internal states.
\item 
The variant without situational awareness achieves the best performance on the two generalization benchmarks, likely because these tasks emphasize self-focused reasoning, such as recognizing one’s own actions and motivations. However, it still underperforms the full model by 1.8 points in \textit{Storyline Quality} and 1.3 points in average score, underscoring the essential role of situational awareness in maintaining coherent and context-aware multi-turn interactions.
\end{itemize}
We further provide a component-level case study of this ablation in Appendix~\ref{app:case4} to qualitatively illustrate the distinct roles of self-awareness and situational awareness.

\subsection{Ablation II: Effects of the Two Implicit Reward Mechanisms}
\label{section:reward_ablation}

To evaluate the impact of the two implicit rewards in CogDual, we run RL with five settings of $\lambda_{\text{ICLG}}$ and $\lambda_{\text{LSA}}$. Table~\ref{tab:reward-ablation} highlights three main findings:
(1) All combinations surpass SFT on out-of-distribution benchmarks. Only the hybrid setting ($\lambda_{\text{ICLG}}{=}0.7,,\lambda_{\text{LSA}}{=}0.3$) improves or maintains all in-domain metrics and yields the highest average, suggesting that balanced causal and semantic rewards optimize both narrative coherence and character fidelity.
(2) Pure LSA ($\lambda_{\text{ICLG}}{=}0$) maximizes Anthropomorphism, showing its strength for persona-centric language, but reduces plot coherence.
(3) Pure ICLG ($\lambda_{\text{LSA}}{=}0$) achieves the best \textit{Storyline Consistency} and \textit{Quality}, indicating its importance for causality and narrative structure.

\subsection{Evaluator Robustness and Human Evaluation}

Relying on a single automatic judge risks evaluator‑specific bias, and
automatic scores alone cannot capture the subtleties of narrative quality.
We therefore conduct two complementary studies whose full protocols and
results are deferred to Appendix~\ref{app:deepseek} and Appendix~\ref{app:human_eval}.

\paragraph{Evaluator Robustness.}
Replacing GPT‑4o with two stylistically distinct scorers, DeepSeek‑v3 and
Gemini‑2.0‑Flash, we re‑run the evaluation for each method. As shown in Appendix \ref{app:deepseek}, CogDual-RL retains the top rank on all metrics under both judges, demonstrating that its gains are not an artifact of a particular evaluator.

\paragraph{Human Evaluation.} We conducted a human evaluation comparing CogDual and related methods on the Llama-3.1-Instruct-8B model. The detailed evaluation protocol and results are presented in Appendix~\ref{app:human_eval}. The results demonstrate that CogDual, leveraging its dual cognition mechanism, more effectively captures the complexity and nuanced emotions of characters, achieving superior performance on subjective metrics.

\noindent
These supplementary results strengthen our main claim: dual cognitive reasoning delivers consistent improvements that are robust to evaluator choice and evident to humans.

\section{Conclusion}
In this paper, we introduce CogDual, a RPLA that incorporates a \textit{cognize-then-respond} reasoning paradigm, aiming to leverage dual cognition for more contextually grounded and psychologically coherent responses. Through reinforcement learning with two proposed general-purpose reward schemes, ICLG and LSA, CogDual further improves upon the supervised fine-tuning baseline. It achieves the best performance among comparable methods on the CoSER benchmark and exhibits strong generalization capabilities on both the Cross-MR and LifeChoice benchmarks.

% \clearpage % 或者使用 \newpage
\section*{Limitations}
Despite the strong empirical performance of CogDual on the CoSER benchmark and its robust generalization across multiple role-playing evaluation tasks, several limitations remain to be addressed in future work. First, due to computational constraints, we have not evaluated the effectiveness of our reinforcement learning approach on larger-scale models such as Llama3.1-70B-Instruct, which may further benefit from the proposed reward design. Second, our current experiments are conducted solely on English datasets, and the model's adaptability to non-English contexts, such as Chinese role-playing scenarios, remains unexplored. Third, in the self-awareness module, we rely on the model to extract previously mentioned memory fragments from the input context, without incorporating an explicit retrieval mechanism to access character-specific memory. This may result in the omission of relevant information.

 % This may result in the omission of relevant information, especially in complex or long-horizon interactions.

\section*{Ethics Statement}
The research conducted in this paper aims to equip RPLAs with cognitive capabilities, enabling them to generate contextually grounded and psychologically coherent responses. Throughout the course of this study, we have adhered rigorously to ethical standards to ensure the integrity and validity of our work. All data used in this research are obtained from publicly available sources, ensuring transparency and reproducibility of our experimental procedures. Furthermore, we have taken careful measures to ensure that our research does not cause harm to any individuals or groups, and we are committed to avoiding any form of deception or misuse of information during the course of this study.

\clearpage % 或者使用 \newpage
\bibliography{references}
\bibstyle{acl_natbib}

\appendix
\clearpage % 或者使用 \newpage
\section{Details about Experiments on CoSER}
\label{appendix:details_on_CoSER}
\paragraph{Constructing Trajectories with Dual Cognition Process}
We first construct trajectories with dual cognition process to fine-tune LLMs for acquiring initial cognitive capabilities, following three principles: 

\begin{itemize}[leftmargin=8pt, topsep=2pt,itemsep=2pt]
\item 

To ensure high-quality cognitive trajectories, we introduce stochastic prompting to improve robustness: during sampling, the LLM is prompted with a 50\% chance to generate structured reasoning from a first-person perspective, and a 50\% chance from a third-person perspective, as details in Appendix~\ref{appendix:CoT Data Construction}. Only the trajectories that pass cognitive field checks are retained. Specifically, each trajectory is represented as a tuple $y = (c, d)$, where the cognitive part $c$ is a structured JSON object composed of two main dimensions, as detailed in Section~\ref{dual_cognitive_frame}.

\item To ensure that the reasoning remains faithful to the narrative context and character identity, we use GPT-4o to verify each trajectory along key cognitive dimensions, filtering out those misaligned with the scenario or character profile. Specifically, it includes the following two steps:

(1) Filtering trajectories that do not meet the cognitive reasoning format: As mentioned in line 848, reasoning trajectories consist of <think> JSON-formatted content </think><answer> response </answer>. We first filter out any trajectories that do not meet this format, as well as those missing the cognitive-related keys specified in the JSON.

(2) Filtering low-quality trajectories using GPT-4o: For trajectories that meet the format requirement from Step 1, we perform an additional fine-grained check. We prompt GPT-4o, combining character information and response, to determine whether the reasoning in each field of the reasoning chain aligns with the character background and consistency of dialogue. For example, in a particular scenario, the original text may show that Claire's next action expresses support for Jamie, but the reasoning may state, "\textit{Claire intends to hinder Jamie by ignoring the medical mysteries.}" We filter out such data that contradicts the character’s behavior. 

\item 
% \textbf{Ensuring Generalizability:}   
To enhance generalization beyond specific narrative styles or configurations, we follow CoSER \cite{coser} and construct role-playing training data using diverse instruction templates, while also varying contextual configurations by randomly including or excluding character profiles, plot summaries, and motivations.

\end{itemize}

\paragraph{Training Data Setup}
For the stage-1 SFT, we use the CoSER dataset \cite{coser} as the meta-level data source, which contains real character interactions extracted from 771 well-known novels, covering 17,966 unique characters. Each instance consists of a plot summary, one or more character profiles, and complete original multi-turn dialogues.

Given the size of the CoSER training set (over 300k instances), we subsample for efficient training. Specifically, we first randomly sample 400 unique characters from it. For each selected character, we collect all associated dialogue scenes.
% \textcolor{red}{rank then random sample? seems weird} 
Then, using GPT-4o, we filter their associated scenes based on cognitive relevance, yielding a subset $\mathcal{D}_{\text{cog}}$ with 38,724 instances. Each instance in $\mathcal{D}_{\text{cog}}$ is used to sample 4 cognitive trajectories, from which high-quality examples are retained following the procedure in \textbf{Constructing Trajectories with Dual Cognition Process}. This results in the final supervised training set $\mathcal{D}_{\text{SFT}}$ with 17,762 examples. 

For the RL stage, we sample 10,000 prompts from the broader cognitively filtered dataset \(\mathcal{D}_{\text{cog}}\), rather than restricting to those used in the supervised set \(\mathcal{D}_{\text{SFT}}\). This design choice enhances training stability and encourages generalization by exposing the policy to both seen and unseen dialogue contexts.

\paragraph{Training Settings}
During stage 1 with SFT, we use a batch size of 64 and set the learning rate to 1e-5. The maximum sequence length is set to 10240, and training is conducted for two epochs. In the RL stage, the batch size is set to 8, and we sample 16 response trajectories for each prompt. The two rewards are weighted at a ratio of 7:3, with the choice of weights based on our ablation study in Section \ref{section:reward_ablation}. The training configuration details for SFT and RL are shown in Table \ref{tab:sft-training-config} and Table \ref{tab:rl-training-config}, respectively.

% \begin{table}[h]
% \centering
% \begin{tabular}{l|c|c|c|c}
% \hline
% \textbf{Model} & \textbf{Learning Rate} & \textbf{Batch Size} & \textbf{Max Length} & \textbf{Training Epochs} \\
% \hline
% Llama3.1-8B-Instruct & 1e-5 & 64 & 10240 & 2 \\
% \mid
% Qwen2.5-7B-Instruct    & 1e-5 & 64 & 10240 & 2 \\
% \hline
% \end{tabular}
% \caption{Training configurations for different instruction-tuned models.}
% \label{tab:training-config}
% \end{table}
\begin{table*}[t]
\centering
\begin{tabular}{l|c|c|c|c}
\hline
\textbf{Model} & \textbf{Learning Rate} & \textbf{Batch Size} & \textbf{Max Length} & \textbf{Training Epochs} \\
\hline
Llama3.1-8B-Instruct & 1e-5 & 64 & 10240 & 2 \\
\midrule
Qwen2.5-7B-Instruct    & 1e-5 & 64 & 10240 & 2 \\
\hline
\end{tabular}
\caption{Training configurations for different instruction-tuned models.}
\label{tab:sft-training-config}
\end{table*}

\begin{table*}[t]
\centering
\small
\setlength{\tabcolsep}{4pt} % 调整列宽的间距
\renewcommand{\arraystretch}{1.2} % 调整行高
\begin{tabular}{l|c|c|c|c|c|c|c|c|c|c}
\hline
\textbf{Model} &
\makecell{\textbf{Learning}\\\textbf{Rate}} &
\makecell{\textbf{Training}\\\textbf{Batch Size}} &
\makecell{\textbf{Forward}\\\textbf{Batch Size}} &
\makecell{\textbf{KL}\\\textbf{Coefficient}} &
\makecell{\textbf{Max}\\\textbf{Length}} &
\makecell{\textbf{Sampling}\\\textbf{Temperature}} &
\makecell{\textbf{Clip}\\\textbf{Range}} &
$\makecell{\boldsymbol{\lambda}\\\boldsymbol{_{\text{ICLG}}}}$ &
$\makecell{\boldsymbol{\lambda}\\\boldsymbol{_{\text{LSA}}}}$ &
\makecell{\textbf{Training}\\\textbf{Steps}} \\
\hline
Llama-3.1-8B-Instruct & 4e-7 & 8 & 128 & 0.001 & 8192 & 0.7 & 0.2 & 0.7 & 0.3 & 120 \\
\midrule
Qwen2.5-7B-Instruct    & 4e-7 & 8 & 128 & 0.001 & 8192 & 0.7 & 0.2 & 0.7 & 0.3 & 120 \\
\hline
\end{tabular}
\caption{Detailed training hyperparameters for reward modeling of instruction-tuned models.}
\label{tab:rl-training-config}
\end{table*}

% \usepackage{makecell} % 导言区添加

% \begin{table*}[t]
% \centering
% \small % 改为 small，而不是缩放整个表格
% \begin{tabular}{l|c|c|c|c|c|c|c|c|c|c}
% \hline
% \textbf{Model} &
% \makecell{\textbf{Learning} \\ \textbf{Rate}} &
% \makecell{\textbf{Training} \\ \textbf{Batch Size}} &
% \makecell{\textbf{Forward} \\ \textbf{Batch Size}} &
% \makecell{\textbf{KL} \\ \textbf{Coefficient}} &
% \makecell{\textbf{Max} \\ \textbf{Length}} &
% \makecell{\textbf{Sampling} \\ \textbf{Temperature}} &
% \makecell{\textbf{Clip} \\ \textbf{Range}} &
% $\makecell{\boldsymbol{\lambda_{\text{ICLG}}}}$ &
% $\makecell{\boldsymbol{\lambda_{\text{LSA}}}}$ &
% \makecell{\textbf{Training} \\ \textbf{Steps}} \\
% \hline
% Llama-3.1-8B-Instruct & 4e-7 & 8 & 128 & 0.001 & 8192 & 0.7 & 0.2 & 0.7 & 0.3 & 120 \\
% \midrule
% Qwen25-7B-Instruct    & 4e-7 & 8 & 128 & 0.001 & 8192 & 0.7 & 0.2 & 0.7 & 0.3 & 120 \\
% \hline
% \end{tabular}
% \caption{Detailed training hyperparameters for reward modeling of instruction-tuned models.}
% \label{tab:reward-training-config}
% \end{table*}

\paragraph{Evaluation Datasets} Since our training data is sourced from CoSER, our main experiments are conducted on the CoSER benchmark. The test set consists of the final 10\% of data from 100 original novels, as well as from 100 additional unseen books. For each test case, LLMs sequentially play different roles based on the given plot and character information. Overall role-play performance is evaluated across multiple dimensions.

\section{The Effect of Evaluator Choice on Model Generalization}
\label{app:deepseek}

To further verify the generalization ability of CogDual, we observe that the default evaluation protocol of the CoSER benchmark adopts GPT-4o as the reference evaluator. Moreover, CogDual-SFT is trained on reasoning chains partially generated by GPT-4o, potentially introducing evaluator bias and yielding evaluation results that disproportionately favor GPT-4o-aligned behavior.

To rule out this potential confound, we incorporate two additional evaluators, Deepseek-v3 and Gemini-2.0-Flash, and re-evaluate all methods on the CoSER benchmark using these models. The full results are presented in Table~\ref{tab:deepseek} and \ref{tab:gemini}. The analysis reveals two key findings:

\begin{itemize}[leftmargin=8pt, topsep=2pt,itemsep=0pt]
\item \textbf{Evaluator Robustness.} Across all evaluators, CogDual consistently outperforms all baselines, demonstrating strong robustness to evaluator choice. Notably, when evaluated using Gemini-2.0-Flash, CogDual maintains a significant performance advantage. This suggests low dependency on evaluator-specific inductive biases.
\item \textbf{RL-based Optimization Advantage.} On all evaluators, CogDual-RL surpasses CogDual-SFT, confirming that the RL enhances generalization and reasoning quality. The improvement is particularly evident in dimensions like \textit{Storyline Consistency} and \textit{Character Fidelity}.
\end{itemize}

\begin{table*}[t]
\centering
\resizebox{\textwidth}{!}{
\begin{tabular}{lcccccc}
\toprule
\textbf{Models} & \textbf{Methods} & \textbf{Storyline Consistency} & \textbf{Anthropomorphism} & \textbf{Character Fidelity} & \textbf{Storyline Quality} & \textbf{Average}\\
\midrule
\rowcolor[gray]{0.9}
\multicolumn{7}{c}{\textbf{Closed-Source LLMs}}\\
GPT-4o & Vanilla & 43.51 & 39.75 & 37.07 & 59.58 & 44.98\\
o1-preview & Vanilla & 44.48 & 40.67 & 37.80 & 59.46 & 45.60\\
\midrule
\rowcolor[gray]{0.9}
\multicolumn{7}{c}{\textbf{Open-Source LLMs}}\\
\multirow{5}{*}{Qwen2.5-7B-Instruct}

  & Vanilla & 38.60 & 29.86 & 27.61 & 37.69 & 33.44\\
  & + LongCoT & 41.01 & 27.04 & 28.81 & 38.86 & 33.93\\
  & + CoSER & 37.93 & \textbf{36.65} & 31.14 & 49.75 & 38.86\\
  \rowcolor{lightblue!100} & + \textbf{CogDual-SFT (ours)} & \underline{44.49} & \underline{32.73} & \underline{34.69} & \underline{49.81} & \underline{40.43}\\
  \rowcolor{lightblue!100} & + \textbf{CogDual-RL (ours)} & \textbf{45.66} & 31.72 & \textbf{36.54} & \textbf{52.84} & \textbf{41.94}\\
\midrule
\multirow{5}{*}{LLaMA3.1-8B-Instruct}
  & Vanilla & 26.08 & 22.86 & 16.20 & 33.88 & 24.76\\
  & + LongCoT & 41.12 & 27.50 & 28.78 & 38.61 & 34.00\\
  & + CoSER & 41.44 & \textbf{38.29} & \underline{34.87} & \underline{54.88} & 42.37\\
  \rowcolor{lightblue!100} & + \textbf{CogDual-SFT (ours)} & \underline{44.88} & \underline{37.95} & 34.27 & \textbf{57.24} & \underline{43.59}\\
  \rowcolor{lightblue!100} & + \textbf{CogDual-RL (ours)} & \textbf{47.94} & 35.89 & \textbf{38.62} & 52.92 & \textbf{43.84}\\
\bottomrule
\end{tabular}}
\caption{Performance of CogDual and baselines on the \textbf{Deepseek-v3}. The best result in each block is in \textbf{bold}; the second-best is \underline{underlined}.}
\label{tab:deepseek}
\end{table*}

\begin{table*}[t]
\centering
\resizebox{\textwidth}{!}{
\begin{tabular}{lcccccc}
\toprule
\textbf{Models} & \textbf{Methods} & \textbf{Storyline Consistency} & \textbf{Anthropomorphism} & \textbf{Character Fidelity} & \textbf{Storyline Quality} & \textbf{Average}\\
\midrule
\rowcolor[gray]{0.9}
\multicolumn{7}{c}{\textbf{Closed-Source LLMs}}\\
GPT-4o & Vanilla & 49.98 & 43.87 & 51.03 & 75.17 & \underline{55.01}\\
GPT-o1-Preview & Vanilla & \textbf{50.42} & \textbf{44.05} & \textbf{54.24} & \textbf{80.97} & \textbf{57.42}\\
\midrule
\rowcolor[gray]{0.9}
\multicolumn{7}{c}{\textbf{Open-Source LLMs}}\\
\multirow{5}{*}{Qwen2.5-7B-Instruct}
  & Vanilla & 50.94 & 33.70 & 42.05 & 47.76 & 43.61\\
  & + LongCoT & 49.19 & 32.84 & 46.75 & 47.90 & 44.17\\
  & + CoSER & 44.96 & \textbf{46.12} & 50.21 & \textbf{71.36} & 53.16\\
  \rowcolor{lightblue!100} & + \textbf{CogDual-SFT (ours)} & \underline{53.40} & \underline{49.72} & \underline{59.08} & 69.31 & \underline{57.88}\\
  \rowcolor{lightblue!100} & + \textbf{CogDual-RL (ours)} & \textbf{56.05} & 47.95 & \textbf{60.69} & \underline{69.63} & \textbf{58.58}\\
\midrule
\multirow{5}{*}{LLaMA3.1-8B-Instruct}
  & Vanilla & 40.62 & 23.59 & 29.08 & 62.74 & 39.01\\
  & + LongCoT & 47.37 & 30.15 & 45.58 & 54.07 & 44.29\\
  & + CoSER & 45.28 & \textbf{48.49} & 47.99 & \underline{74.91} & 54.17\\
  \rowcolor{lightblue!100} & + \textbf{CogDual-SFT (ours)} & \underline{48.95} & 46.36 & \underline{50.97} & \textbf{75.78} & \underline{55.51}\\
  \rowcolor{lightblue!100} & + \textbf{CogDual-RL (ours)} & \textbf{54.93} & \underline{46.50} & \textbf{55.55} & 68.99 & \textbf{56.49}\\
\bottomrule
\end{tabular}}
\caption{Performance of CogDual and baselines evaluated with \textbf{Gemini-2.0-Flash}. The best result in each block is in \textbf{bold}; the second-best is \underline{underlined}.}
\label{tab:gemini}
\end{table*}

\section{The Details of Human Evaluation}
\label{app:human_eval}
To more objectively assess subjective qualities, we conducted additional experiments involving human evaluation to provide a balanced and credible assessment. Specifically, we selected 50 samples from classic works in the CoSER test set, including \textit{Pride and Prejudice, A Game of Thrones, War and Peace, Les Misérables,} and \textit{The Complete Sherlock Holmes}, and generated dialogues using LLaMA3.1-8B-Instruct. Five evaluators familiar with these characters assessed the responses produced by Vanilla, CoSER, LongCoT, CogDual-SFT, and CogDual-RL models across four dimensions:\textit{ Storyline Consistency}, \textit{Anthropomorphism}, \textit{Character Fidelity}, and \textit{Storyline Quality}. For each case, the evaluators selected the best response. We report the Win Rate for each model, as shown in the Table \ref{tab:human-eval}.

\begin{table*}[!t]
\centering
\large
\resizebox{\textwidth}{!}{
\begin{tabular}{l|l|ccccc}
\toprule
\textbf{Model} & \textbf{Method} & \textbf{Storyline Consistency} & \textbf{Anthropomorphism} & \textbf{Character Fidelity} & \textbf{Storyline Quality} & \textbf{Win Rate (Avg.)} \\
\midrule
\multirow{5}{*}{\centering LLaMA3.1-8B-Instruct}
    & Vanilla              & 2.80  & 2.00  & 4.80  & 3.20  & 3.20   \\
    & +CoSER               & 21.60 & 30.80 & 28.00 & 26.40 & 26.70  \\
    & +LongCoT             & 17.20 & 4.00  & 7.60  & 8.00  & 9.20   \\
    & +CogDual-SFT (ours)  & 27.20 & 30.80 & \textbf{31.60} & 29.60 & 29.80  \\
    & +CogDual-RL (ours)   & \textbf{31.20} & \textbf{32.40} & 28.00 & \textbf{32.80} & \textbf{31.10} \\
\bottomrule
\end{tabular}
}
\caption{
Human evaluation win rates (\%) across four dimensions for different methods based on \texttt{LLaMA3.1-8B-Instruct} on the CoSER test set. The highest value for each metric is highlighted in \textbf{bold}.
}
\label{tab:human-eval}
\end{table*}

\section{A Reference Prompt for the CoT Data Construction}
\label{appendix:CoT Data Construction}
\begin{table*}[!h]
    \centering
    \small
    \begin{tabular}{p{16cm}}  % Adjust the width to span both columns
    \toprule
     \rowcolor[gray]{0.9} 
     \multicolumn{1}{c}{\textbf{A Reference Prompt for the CoT Data Construction}} \\
     \midrule

You are a role play expert. Your task is to generate the \textbf{reasoning process} of \{character\} before the action step by step, based on the character's profile, scene context, and the historical dialogues of multiple characters from the current situation. You should output the reasoning process using \texttt{<think>} and \texttt{</think>} to wrap the reasoning process. \\

\textbf{\#\# Current Input of the Situational Interpretation Information} \\
\\
\textbf{=== Character Played ===} \\
\{character\} \\
\\

\textbf{=== \{character\}'s Profile ===} \\
\{character\_profile\} \\
\\

\textbf{=== Other Characters in the Scene ===} \\
\{other\_characters\_profile\} \\
\\

\textbf{=== Current Scenario ===} \\
\{current\_scenario\} \\
\\

\textbf{=== \{character\}'s Psychological or Motivational State in the Scene ===} \\
\{thought\} \\
\\

\textbf{=== Historical Dialogue in the Current Situation ===} \\
\{history\_str\} \\
\\

\textbf{=== \{character\}'s Next [thought], (action), speech ===} \\
\{assistant\_content\} \\
\\

\textbf{\#\# Attention} \\
- The reasoning process you output is actually the character's analysis before making the Next [thought], (action), speech. \\
\\

\textbf{\#\# Output Format\{use\_first\_person\}} \\
\texttt{<think>} \\
your reasoning process... \\
\texttt{</think>} \\
So that the next [thought], (action), speech of \{character\} could be: \\
\texttt{<answer>} \\
\{assistant\_content\} \\
\texttt{</answer>} \\

    \bottomrule
    \end{tabular}
    \caption{A reference prompt for the CoT data construction.}
    \label{table:cot_construction}
\end{table*}
As described in Section~\ref{baselines}, we construct the LongCoT baseline using the prompt shown in Table~\ref{table:cot_construction}.

\section{A Reference Prompt for Constructing CogDual Training Data}
\begin{table*}[!h]
    \centering
    \scriptsize
    \begin{tabular}{p{16cm}}  % Adjust the width to span both columns
    \toprule
     \rowcolor[gray]{0.9} 
     \multicolumn{1}{c}{\textbf{A Reference Prompt for Generating Dual Cognitive Reasoning before Character Responses}} \\
     \midrule
You are a psychology expert with deep knowledge of cognitive behaviors. Your task is to generate the cognitive reasoning process of \{character\} before the action, based on the definition of dual cognition, and by integrating the character's profile, scene context, and the historical dialogues of multiple characters from the current situation. \\

Please follow the definition of cognitive behavior provided below to simulate \{character\}'s psychological state, motivations, and analysis of the environment/others. Focus specifically on how the reasoning process influences \{character\}'s upcoming response in the plot. \\

\textbf{\#\# Dual Cognitive Psychology Definition of the Character} \\

The dual cognitive process unfolds from the external environment to the internal self. First, \{character\} assesses the current situation based on their identity, quickly making judgments about the context. Next, based on these judgments, \{character\} analyzes the behavior and speech of others to infer their intentions and the overall scene context. This analysis leads to self-awareness, where \{character\} identifies their emotional state, motivations, and focus in the given context. Finally, based on all these perceptions, \{character\} forms a cognitive strategy and psychological activity before moving forward with the next action in the plot. \\
\\
\textbf{\#\# Dual Cognitive Reasoning Process} \\

The reasoning steps of dual cognition primarily include two parts: situational awareness analysis and self-awareness analysis, as outlined below. \\

\textbf{1. Situational Awareness Analysis} \\
- \textbf{Situation Perception}: Which aspects of the current situation—such as environmental factors, changes in events, or immediate challenges—could influence \{character\}'s emotions, thoughts, or decisions in the near future? \\
- \textbf{Perception of Others}: This includes interpreting the behaviors, emotional states, and potential intentions of other characters present in the scene. \\
- \textbf{Behavior Analysis}: Considering both the current scene and historical dialogues, which actions or words from others might be noteworthy and could influence \{character\}'s response? \\
- \textbf{Emotion Analysis}: Based on the current situation and the behavior of others, what emotions might \{character\} perceive from others? How could these emotions affect \{character\}? \\
- \textbf{Intentions Analysis}: In light of the situation and the behaviors and emotions of others, what could be the explicit or implicit intentions behind others’ actions? How might \{character\} perceive these intentions? \\

\textbf{2. Self-Awareness Analysis} \\
- \textbf{Key Memory Activation}: Based on the situational awareness, what past experiences or memories of \{character\} might be triggered by the current situation? Which specific memories could influence \{character\}'s response? \\
- \textbf{Self-Emotion}: Based on the situational and behavioral analysis, what emotions is \{character\} currently experiencing? For example, are they feeling doubt, hope, anxiety, or fear? How do these emotions relate to the unfolding situation? \\
- \textbf{Self-Intentions}: Based on the emotional and situational analysis, what are \{character\}'s primary motivations or intentions at this moment? How do these intentions shape their decision-making? \\
- \textbf{Internal Thoughts and Strategy}: Drawing from all of the above—background, situational awareness, and self-awareness—what are \{character\}'s internal thought processes and strategies? How does \{character\} plan to proceed, and what cognitive steps are taken before executing next thought, action, speech? \\
\\
\textbf{\#\# Current Input of the Situational Interpretation Information} \\
=== Character Played === \\
\{character\} \\

=== \{character\}'s Profile === \\
\{character\_profile\} \\

=== Other Characters in the Scene === \\
\{other\_characters\_profile\} \\

=== Current Scene Description === \\
\{current\_scenario\} \\

=== \{character\}'s Psychological or Motivational State in the Scene === \\
\{thought\} \\

=== Historical Dialogue in the Current Situation === \\
\{history\_str\} \\

=== \{character\}'s Next [thought],(action),speech === \\
\{assistant\_content\} \\

\textbf{\#\# Attention} \\
- The cognitive reasoning you output is actually the character's analysis before making the Next [thought], (action), speech. \\
- For each cognitive dimension, you only need to grasp the key points for analysis. The content between dimensions should be continuous, with a hierarchical logic and as little repetition as possible. (for example, gradually transitioning from situational awareness to deep self-awareness) \\
\textbf{\#\# Output Format\{use\_first\_person\}} \\
First, I need to simulate \{character\}'s cognitive process briefly before the next [thought],(action),speech. \\
\begin{minipage}{\linewidth}
\begin{verbatim}
<coginitive>
{{
  "situational_awareness": {{
    "environmental_perception": "...",
    "others_perception": {{
      "behavior": {{
        "character1": "...",
        ...  
      }},
      "emotion": {{  
        "character1": "this character's emotions", 
        ...  
      }},
      "intentions": {{  
        "character1": "inferred intention1",
        ...  
      }}
    }}
  }},  
  "self_awareness": {{
    "key_memory": ["memories relevant to the current situation"], 
    "current_emotions": "...",  
    "perceived_intentions": "...", 
    "internal_thought": "..."
  }}
}}
</coginitive>
So that the next [thought],(action),speech of {character} could be:
<answer>
{assistant_content}
</answer>
\end{verbatim}
\end{minipage}
\\
     \bottomrule
    \end{tabular}
    \caption{A Reference Prompt for Generating Dual Cognitive Reasoning before Character Responses.}
    \label{table:cogdual_data}
\end{table*}
As described in \textbf{Constructing Trajectories with Dual Cognition Process}, we use the prompt in Table~\ref{table:cogdual_data} to generate cognitive reasoning trajectories with GPT-4o.

\begin{table*}[!h]
    \centering
    \small
    \begin{tabular}{p{16cm}}  % Adjust the width to span both columns
    \toprule
    \rowcolor[gray]{0.9} 
    \multicolumn{1}{c}{\textbf{A Reference Prompt for Filtering Subset.}} \\
    \midrule
You are a cognitive behavior analyst tasked with determining which of the character's dialogues in a given scenario require the generation of cognitive reasoning (as defined below). Your goal is to filter dialogues where the character demonstrates situational awareness (environmental/others perception) or self-awareness (memory, motivation, emotion, internal state), and flag them as needing cognitive reasoning.

\textbf{\#\#\# Cognitive Behavior Definition} \\
Cognitive reasoning is required for dialogues where the character exhibits:

\textbf{\#\#\# Situational Awareness:} \\
- Environmental Perception: Notice of environmental details affecting behavior (e.g., "The dim lighting made her hesitate"). \\
- Others Perception: Inference about others' intentions, emotions, or behavior patterns (e.g., "Her calm tone suggested she was hiding something"). \\

\textbf{\#\#\# Self-Awareness:} \\
- Memory Activation: Reference to past events influencing current actions (e.g., "This room reminded him of his childhood home"). \\
- Motivations: Clear prioritization of goals (e.g., "I need to confirm her loyalty before sharing secrets"). \\
- Current Emotions: Recognition of emotional states affecting behavior (e.g., "Anger clouded his judgment, so he paused"). \\
- Internal State: Awareness of cognitive/mental state (e.g., "Fatigue made it hard to focus, but he pressed on"). \\

\textbf{\#\# Task Instructions} \\
\textbf{\#\#\# Parse the Dialogue:} \\
- Split the dialogue into turns, focusing on the character's lines (e.g., "Robert Neville: [thought] response"). \\

\textbf{\#\#\# Identify Cognitive Triggers:} For each of the \{character\}'s lines, check if: \\
- The bracketed thought (if present) explicitly describes situational/self-awareness (use the definition above). \\
- The spoken response implicitly requires reasoning about environment, others, or self (even without explicit thoughts, e.g., a question that reflects suspicion of others' motives). \\

\textbf{\#\#\# Filter Criteria:} \\
- Need Cognitive Reasoning: Dialogue turns where the \{character\}'s thought/response involves analysis of environment, others' behavior, personal motivations, or emotions (as in the example below). \\
- No Cognitive Reasoning Needed: Simple actions (e.g., "nods silently"), neutral responses (e.g., "Yes"), or dialogues lacking explicit/implicit awareness of the cognitive components above. \\

\textbf{\#\# Output Format:} \\
List each dialogue turn that needs cognitive reasoning, with a brief reason, like: \\
\begin{verbatim}
[
    {
      "index": 0,
      "needs_cognitive": (true or false),
      "reason": ...
    },
    ...
]
\end{verbatim}

\textbf{\#\# Example} \\
\textbf{\#\#\# Input Example} \\
\{input\_example\}

\textbf{\#\#\# Output Example} \\
\{output\_example\}
    \\
    \bottomrule
    \end{tabular}
    \caption{A reference prompt for filtering subset.}
    \label{table:filter}
\end{table*}
\begin{table*}[!h]
    \centering
    \small
    \begin{tabular}{p{16cm}}  % 单栏控制宽度
    \toprule
    \rowcolor[gray]{0.9} 
    \multicolumn{1}{c}{\textbf{A reference prompt used for CoT Prompting}} \\
    \midrule
    \begin{minipage}[t]{\linewidth}
    \ttfamily
You are \{character\} from \{book\_name\}.\\

===\{character\}'s Profile===\\
\{character\_profile\}\\

===Current Scenario===\\
\{scenario\}\\
\{other\_character\_profiles\_str\}\{motivation\}\\

===Requirements===\\
Your output should include \textbf{think}, \textbf{thought}, \textbf{speech}, and \textbf{action}. Before responding, first think using <think> tags:\\

<think>your thinking</think>\\

After your thinking, your output should include \textbf{thought}, \textbf{speech}, and \textbf{action}.\\
Use [your thought] for thoughts, which others can't see.\\
Use (your action) for actions, which others can see.\\

===Output Example===\\
\{REASONING\_EXAMPLE\}\\

===Your Output=== (let's think step by step!)\\
    \end{minipage} \\
    \bottomrule
    \end{tabular}
    \caption{A reference prompt used for CoT Prompting.}
    \label{table:cot_system_prompt}
\end{table*}

\section{A Reference Prompt for Cognitive-Based Chain-of-Thought}
\label{appendix:cb-cot}

To further validate the effectiveness of our dual cognition framework, we propose a low-cost and cognitive-based Chain-of-Thought prompting approach(CB-CoT). Specifically, the LLM is guided to understand dual-cognition reasoning through in-context definitions and instructed to produce outputs in the same structured format as CogDual in Section~\ref{dual_cognitive_frame}. The whole prompt design is shown in Table~\ref{table:cb_cot_method}.

\label{appendix:cb_cot}
\begin{table*}[!h]
    \centering
    \small
    \begin{tabular}{p{16cm}}  % Adjust the width to span both columns
    \toprule
    \rowcolor[gray]{0.9} 
    \multicolumn{1}{c}{\textbf{A Reference Prompt for Cognitive-Based Chain-of-Thought}} \\
    \midrule
\texttt{You are \{character\} from \{book\_name\}.}

\vspace{1mm}
\\
\texttt{===\{character\}'s Profile===} \\
\texttt{\{character\_profile\}}
\\
\vspace{1mm}
\texttt{===Current Scenario===} \\
\texttt{\{scenario\}} \\
\texttt{\{other\_character\_profiles\_str\}} \\
\texttt{\{motivation\}}
\\
\vspace{1mm}
\texttt{===Requirements===} \\
Your output should include \textbf{cognitive think}, \textbf{thought}, \textbf{speech}, and \textbf{action}. Before responding, first use \texttt{<think>} tags for your cognitive analysis like human thought, which others cannot see:
\\
\texttt{\{cognition\_ process\}}
\\
\\
\texttt{<think>} \\
\texttt{\{} \\
\hspace*{3mm}\texttt{"situational\_awareness": \{} \\
\hspace*{6mm}\texttt{"environmental\_perception": "..."}, \\
\hspace*{6mm}\texttt{"others\_perception": \{} \\
\hspace*{9mm}\texttt{"behavior": \{} \\
\hspace*{12mm}\texttt{"character1": "..."}, \\
\hspace*{12mm}\texttt{...} \\
\hspace*{9mm}\texttt{\}}, \\
\hspace*{9mm}\texttt{"emotion": \{} \\
\hspace*{12mm}\texttt{"character1": "this character's emotions"}, \\
\hspace*{12mm}\texttt{...} \\
\hspace*{9mm}\texttt{\}}, \\
\hspace*{9mm}\texttt{"intentions": \{} \\
\hspace*{12mm}\texttt{"character1": "inferred intention1"}, \\
\hspace*{12mm}\texttt{...} \\
\hspace*{9mm}\texttt{\}} \\
\hspace*{6mm}\texttt{\}}, \\
\hspace*{3mm}\texttt{\}}, \\
\hspace*{3mm}\texttt{"self\_awareness": \{} \\
\hspace*{6mm}\texttt{"key\_memory": ["memories relevant to the current situation"]}, \\
\hspace*{6mm}\texttt{"current\_emotions": "..."}, \\
\hspace*{6mm}\texttt{"perceived\_intentions": "..."}, \\
\hspace*{6mm}\texttt{"internal\_thought": "..."} \\
\hspace*{3mm}\texttt{\}} \\
\texttt{\}} \\
\texttt{</think>}

[your thought] \\
your speech \\
(your action)

\vspace{1mm}
\texttt{===Your Output===}
    \\
    \bottomrule
    \end{tabular}
    \caption{A reference prompt used for generating dual cognition reasoning(CB-CoT) before character responses. The \textit{cognition process} is detailed in Table \ref{table:cognition_process}}.
    \label{table:cb_cot_method}
\end{table*}

\begin{table*}[!h]
    \centering
    \small
    \begin{tabular}{p{16cm}}  % Span both columns
    \toprule
    \rowcolor[gray]{0.9} 
    \multicolumn{1}{c}{\textbf{The Definition of the Cognition Process}} \\
    \midrule

\textbf{1. Situational Awareness Analysis} \\
\textbf{Situation Perception:} Which aspects of the current situation—such as environmental factors, changes in events, or immediate challenges—could influence \{character\}'s emotions, thoughts, or decisions in the near future? \\

\textbf{Perception of Others:} Interpreting the behaviors, emotional states, and potential intentions of other characters present in the scene. \\

\textbf{Behavior Analysis:} Considering both the current scene and historical dialogues, which actions or words from others might be noteworthy and could influence \{character\}'s response? \\

\textbf{Emotion Analysis:} Based on the current situation and the behavior of others, what emotions might \{character\} perceive? How could these emotions affect them? \\

\textbf{• Intentions Analysis:} In light of the situation and the behaviors and emotions of others, what are the explicit or implicit intentions behind others’ actions? \\

\vspace{1mm}
\textbf{2. Self-Awareness Analysis} \\

\textbf{Key Memory Activation:} What past experiences or memories might be triggered by the current situation? Which specific memories could influence \{character\}'s response? \\

\textbf{Self-Emotion:} What emotions is \{character\} currently experiencing (e.g., doubt, hope, anxiety)? How do these emotions relate to the current situation? \\

\textbf{Self-Intentions:} What are \{character\}'s primary motivations or goals at this moment? How do they shape decision-making? \\

\textbf{Internal Thoughts and Strategy:} Based on all of the above, what are \{character\}'s internal thought processes? What strategy guides their next action, thought, or speech? \\

    \bottomrule
    \end{tabular}
    \caption{The definition of the cognition process.}
    \label{table:cognition_process}
\end{table*}

\section{A Reference Prompt for Semantic Matching}
\label{appendix:semantic_matching}
We use GPT-4o to choose the option that is most semantically similar to the response part generated by CogDual. The prompt is shown in Table \ref{table:similar_semantic}
\begin{table*}[!h]
    \centering
    \small
    \begin{tabular}{p{16cm}}  % Adjust the width to span both columns
    \toprule
    \rowcolor[gray]{0.9} 
    \multicolumn{1}{c}{\textbf{A Reference Prompt for Semantic Matching}} \\
    \midrule
Please select the option among the following four sentences that is semantically closest to the \texttt{target\_sentence}.\\

\vspace{0.5em}
Options: \texttt{\{options\}}\\

Target sentence: \texttt{\{target\_sentence\}}\\

\vspace{0.5em}
Your output should be structured as the following schema:\\
\texttt{\{"Choice": str // "A"/"B"/"C"/"D", "Reason": string // The reason of the choice\}}\\
    \bottomrule
    \end{tabular}
    \caption{A reference prompt for semantic matching}
    \label{table:similar_semantic}
\end{table*}

\section{Case Study}
We select two representative CogDual reasoning cases from the test set to analyze the effectiveness and granularity of the model’s cognitive reasoning process.
\subsection{Case 1: Catherine Leigh Dollanganger in Flowers in the Attic}
Tables~\ref{tab:case1_1} and~\ref{tab:case1_2} showcase a representative scenario from Flowers in the Attic (Dollanganger, \#1) and the corresponding simulation by CogDual-RL. In this case, Catherine Leigh Dollanganger, faced with Bart’s disappointment and emotional turmoil during the failed Christmas ball, delivers a gentle yet honest response that balances empathy with practical advice.

This outcome reflects the model’s ability to accurately capture and reason about both the external environment and internal motivations, as seen in the detailed dual cognition trace. The simulation not only recognizes Bart’s visible frustration but also draws on Catherine’s personal memories and sense of responsibility, resulting in an action that is deeply aligned with her character and the unfolding narrative context.

Such behavior demonstrates that CogDual-RL can produce responses that are both contextually appropriate and consistent with character persona, indicating effective integration of narrative knowledge and cognitive reasoning within the model.

\renewcommand{\arraystretch}{1.4}

\begin{table*}[!ht]
\centering
\small
\begin{tabularx}{\linewidth}{>{\raggedright\arraybackslash}p{3cm} X}
\toprule
\multicolumn{2}{c}{\textbf{Flowers in the Attic (Dollanganger, \#1)} \hfill (From Test Set)} \\
\midrule

\textbf{Character} & \textbf{Catherine Leigh Dollanganger} \\
\midrule

\textbf{Topic} & The failed Christmas ball and its aftermath \\
\midrule

\textbf{Plot Summary} & 
Bart hosts a grand Christmas ball at Foxworth Hall, but none of the invited guests show up. The family tries to salvage the evening, with Cindy entertaining everyone with her singing. Bart becomes increasingly angry and bitter, vowing revenge on those who snubbed him. Cathy suspects Joel of sabotaging the party by not mailing the invitations and confronts him. \\
\midrule

\textbf{Scenario} & 
It's Christmas Eve at Foxworth Hall, a grand mansion decorated lavishly for a highly anticipated ball. The enormous ballroom sparkles with ornate decorations, a towering Christmas tree, and tables laden with gourmet food and champagne. The Foxworth family, dressed in their finest attire, anxiously awaits the arrival of their distinguished guests. As the appointed hour passes with no arrivals, tension mounts... \\
\midrule

\textbf{Character Profile} & 
\textbf{Bart Winslow} \newline
Bart is devastated when no guests arrive for his grand ball. He becomes increasingly bitter and angry, vowing revenge on those who snubbed him. His vulnerability and hurt are evident beneath his rage... \newline

\textbf{Catherine Leigh Dollanganger} \newline
Cathy tries to comfort Bart and salvage the evening. She suspects Joel of sabotaging the party and confronts him about it. She's deeply concerned about Bart's emotional state and vengeful plans... \newline

\textbf{Cindy Sheffield} \newline
Cindy surprises everyone with her beautiful singing voice, entertaining the family and staff during the failed party. She tries to cheer up Bart and make the best of the situation... \\
\midrule

\textbf{History Dialogue} & 
\textbf{Bart Winslow:} [I can't believe this is happening. The failure of the ball is a personal affront.] (steps forward, his voice carrying a mix of frustration and determination) I'm sorry, but it seems we've been let down. I suppose it's up to me to salvage the situation. (gestures around the empty ballroom, his eyes narrowing with a sense of challenge) \newline

\textbf{Catherine Leigh Dollanganger:} [I can see the disappointment in Bart's eyes, and it breaks my heart. This is yet another failure in a long line of them.] Bart, dear, perhaps we should just call it off for tonight and reschedule for another time. (places a reassuring hand on his arm, hoping to offer some comfort) \newline

\textbf{Bart Winslow:} [I can't let this failure define me. I need to prove my worth.] (shakes his head, determination hardening his features) No, Mother, I won't let this ruin the evening. I'll find a way to make it happen. (turns to the orchestra, his voice firm and resolute) \newline \\
\bottomrule
\end{tabularx}
\caption{An example of dual cognition reasoning for Catherine Leigh Dollanganger in \textit{Flowers in the Attic}.}
\label{tab:case1_1}
\end{table*}

\begin{table*}[!ht]
\centering
\small
\begin{tabularx}{\linewidth}{>{\raggedright\arraybackslash}p{3cm} X}
\toprule
\multicolumn{2}{c}{\textbf{Flowers in the Attic (Dollanganger, \#1)} \hfill (From Test Set)} \\
\midrule

\textbf{Character} & \textbf{Catherine Leigh Dollanganger} \\
\midrule

\textbf{Dual Cognition Reasoning} & 
\begin{minipage}[t]{\linewidth}
\ttfamily
\noindent
\hspace*{0pt}\{\\
\hspace*{1em}"situational\_awareness": \{\\
\hspace*{2em}"environmental\_perception": "The grandeur of Foxworth Hall is starkly contrasted by the emptiness of the ballroom, creating a sense of disappointment and urgency as the Christmas Eve event fails to materialize.",\\
\hspace*{2em}"others\_perception": \{\\
\hspace*{3em}"behavior": \{\\
\hspace*{4em}"Bart Winslow": "Bart steps forward with a mix of frustration and determination, indicating his willingness to take charge of the situation despite the chaos."\\
\hspace*{3em}\},\\
\hspace*{3em}"emotion": \{\\
\hspace*{4em}"Bart Winslow": "Bart's emotions are a blend of frustration and determination, reflecting his disappointment and resolve to salvage the event."\\
\hspace*{3em}\},\\
\hspace*{3em}"intentions": \{\\
\hspace*{4em}"Bart Winslow": "Bart's intention is to salvage the situation, driven by a sense of responsibility and ambition to uphold the Foxworth legacy."\\
\hspace*{3em}\}\\
\hspace*{2em}\}\\
\hspace*{1em}\},\\
\hspace*{1em}"self\_awareness": \{\\
\hspace*{2em}"key\_memory": [\\
\hspace*{3em}"The memory of past disappointments and failures in the family, particularly related to Bart's ambitions and the Foxworth legacy."\\
\hspace*{2em}],\\
\hspace*{2em}"current\_emotions": "A mix of disappointment and concern, worried about the impact of this failure on Bart and the family's reputation.",\\
\hspace*{2em}"perceived\_intentions": "I need to support Bart and help him navigate this crisis, ensuring the family's reputation remains intact.",\\
\hspace*{2em}"internal\_thought": "I need to reassure Bart and offer him guidance, drawing from my own experiences of dealing with family failures and disappointments."\\
\hspace*{1em}\}\\
\hspace*{0pt}\}
\end{minipage}
\\
\midrule

\textbf{Simulation response from CogDual-RL} & 
\textit{[I see the fire in Bart’s eyes, and it breaks my heart—he’s trying so hard to hold it together.]} Bart, darling... I know how much this meant to you. But maybe... maybe it’s time to let the night go. (She gently places a hand on his arm.) Pushing through might only hurt you more. There’s no shame in pausing. \\
\bottomrule
\end{tabularx}
\caption{An example of dual cognition reasoning for Catherine Leigh Dollanganger in \textit{Flowers in the Attic}.}
\label{tab:case1_2}
\end{table*}

\subsection{Case 2: Nicholas of Morimondo in The Name of the Rose}
Tables~\ref{tab:case2_1} and~\ref{tab:case2_2} present a representative example from The Name of the Rose, showcasing the dual cognition reasoning and simulated response for Nicholas of Morimondo. Table~\ref{tab:case2_1} sets the scene—a tense exchange in the abbey’s crypt, rich with historical and political undercurrents—while Table~\ref{tab:case2_2} displays the detailed cognitive reasoning trace and the corresponding output from CogDual-RL.

The reasoning trace reveals that Nicholas is acutely aware of both the sacred atmosphere of the crypt and the probing intentions of William. He draws upon memories of past interrogations, balancing his pride in the abbey’s legacy with caution and a desire to protect institutional secrets. This nuanced internal process leads directly to his simulated reply: Nicholas offers a measured, carefully worded answer that acknowledges the political importance of the librarian position without revealing sensitive details.

This example demonstrates CogDual’s ability to generate in-character responses grounded in a fine-grained cognitive process, effectively integrating environmental cues, social context, and personal motivation. The clear causal link between Nicholas’s internal reasoning and his speech highlights the model’s strengths in both contextual fidelity and interoperability.

\renewcommand{\arraystretch}{1.4}

\begin{table*}[!ht]
\centering
\small
\begin{tabularx}{\linewidth}{>{\raggedright\arraybackslash}p{3cm} X}
\toprule
\multicolumn{2}{c}{\textbf{The Name of the Rose \#1} \hfill (From Test Set) } \\
\midrule

\textbf{Character} & \textbf{Nicholas of Morimondo} \\
\midrule

\textbf{Topic} & The abbey's treasures and its recent history. \\
\midrule

\textbf{Scenario} & 
The treasure crypt of the abbey is a cavernous, dimly lit chamber filled with an impressive collection of religious artifacts, precious metals, and rare manuscripts. Torchlight flickers against glass cases and ornate display stands, casting long shadows across the stone walls. The air is heavy with the musty scent of ancient objects and the weight of centuries of monastic history. Shelves and cabinets line the walls, each holding relics that represent the abbey's immense wealth and spiritual significance. William, Adso, and Nicholas move carefully through this sacred space, their footsteps echoing softly against the stone floor. \\
\midrule

\textbf{Character Profile} & 
\textbf{Nicholas of Morimondo} \newline
Nicholas guides William and Adso through the treasure crypt, showing them relics and sharing information about the abbey's recent history and politics... \newline

\textbf{William of Baskerville} \newline
William questions Nicholas about Malachi and the history of the librarian position, trying to gather information relevant to his investigation... \newline

\textbf{Adso of Melk} \newline
Adso is amazed by the treasures in the crypt and listens attentively to the conversation between William and Nicholas... \\
\midrule

\textbf{History Dialogue} & 
\textbf{Nicholas of Morimondo:} [I'll start with something impressive.] (gestures towards a case) This evangeliarium has a binding with amazing enamel plaques and precious stones. \newline

\textbf{William of Baskerville:} [The treasures are fascinating, but I need information.] These are indeed wondrous objects, Nicholas. Tell me, what sort of man was Malachi? \newline

\textbf{Nicholas of Morimondo:} [I must be careful with my words.] (hesitates) Malachi seemed quite thoughtful, but on the contrary, he was a very simple man. According to Alinardo, he was a fool. \newline 

\textbf{William of Baskerville:} [Interesting. There's more to this.] Alinardo bears a grudge against someone for a remote event, when he was denied the honor of being librarian, doesn't he? \newline

\textbf{Nicholas of Morimondo:} [He's perceptive. I should provide more context.] Indeed. It's an old story, dating back at least fifty years. There have always been whispers about the library and its keepers. \newline

\textbf{Adso of Melk:}  [This is intriguing.] (looks at William) Master, why is the position of librarian so important? \newline 

\textbf{William of Baskerville:}  [Good question, Adso.] Nicholas, can you explain the significance of the librarian's position in the abbey? \newline 
\\

\bottomrule
\end{tabularx}
\caption{An example of dual cognition reasoning for Nicholas of Morimondo in \textit{The Name of the Rose}.}
\label{tab:case2_1}
\end{table*}

\begin{table*}[!ht]
\centering
\small
\begin{tabularx}{\linewidth}{>{\raggedright\arraybackslash}p{3cm} X}
\toprule
\multicolumn{2}{c}{\textbf{The Name of the Rose \#2} \hfill (From Test Set)} \\
\midrule

\textbf{Character} & \textbf{Nicholas of Morimondo} \\
\midrule

\textbf{Dual Cognition Reasoning} & 
\begin{minipage}[t]{\linewidth}
\ttfamily
\noindent
\hspace*{0pt}\{\\
\hspace*{1em}"situational\_awareness": \{\\
\hspace*{2em}"environmental\_perception": "The crypt, steeped in age and silence, radiates reverence. It is a place meant to preserve, not reveal.",\\
\hspace*{2em}"others\_perception": \{\\
\hspace*{3em}"behavior": \{\\
\hspace*{4em}"William of Baskerville": "William observes everything with clear investigative purpose. His questions are well-timed, deliberate, and increasingly penetrating.",\\
\hspace*{4em}"Adso of Melk": "Adso follows William closely, occasionally reacting with surprise or reverence."\\
\hspace*{3em}\},\\
\hspace*{3em}"emotion": \{\\
\hspace*{4em}"William of Baskerville": "William is composed but vigilant.",\\
\hspace*{4em}"Adso of Melk": "Adso seems curious and trusting, possibly still naive about the politics underlying their inquiry."\\
\hspace*{3em}\},\\
\hspace*{3em}"intentions": \{\\
\hspace*{4em}"William of Baskerville": "William is probing for political significance masked as historical detail.",\\
\hspace*{4em}"Adso of Melk": "Adso wants to learn and support William, but may also be searching for meaning that reconciles reason with faith."\\
\hspace*{3em}\}\\
\hspace*{2em}\}\\
\hspace*{1em}\},\\
\hspace*{1em}"self\_awareness": \{\\
\hspace*{2em}"key\_memory": [\\
\hspace*{3em}"Nicholas recalls past visits from external investigators or inquisitors where a single misworded answer led to suspicion and scrutiny."\\
\hspace*{2em}],\\
\hspace*{2em}"current\_emotions": "Nicholas feels a mix of pride in the abbey's history and caution in sharing sensitive information.",\\
\hspace*{2em}"perceived\_intentions": "Remain helpful and cooperative on the surface, while ensuring that William and Adso do not uncover anything that could compromise the institution.",\\
\hspace*{2em}"internal\_thought": "William's questions are circling too close to the truth. I must answer without lying, but I cannot give him everything."\\
\hspace*{1em}\}\\
\hspace*{0pt}\}
\end{minipage}
\\
\midrule

\textbf{Simulation response from CogDual-RL} & 
\textit{[This is a delicate balance. I must guide them without revealing too much.]} (clears throat, voice measured) The librarian holds not just keys to books, but to knowledge and power. By longstanding tradition, the role often leads to becoming abbot. \\
\bottomrule
\end{tabularx}
\caption{An example of dual cognition reasoning for Nicholas of Morimondo in \textit{The Name of the Rose}.}
\label{tab:case2_2}
\end{table*}

\subsection{Case 3: An example of an extracted conversation and its multi-agent simulation}
We present a simulation from The Dragon Reborn to evaluate CogDual’s effectiveness, as shown in Tables~\ref{tab:case3_1} through~\ref{tab:case3_6}. The dialogue shows that CogDual captures both Perrin’s internal struggle to appear strong and the supporting characters’ distinctive reasoning and emotional roles. Each character’s internal thoughts are closely tied to their outward actions, resulting in interactions that are both believable and faithful to the narrative. This demonstrates CogDual’s strength in producing contextually appropriate, character-consistent, and psychologically plausible role-play compared to standard baselines.

\renewcommand{\arraystretch}{1.4}

\begin{table*}[!ht]
\centering
\small
\begin{tabularx}{\linewidth}{>{\raggedright\arraybackslash}p{3cm} X}
\toprule
\multicolumn{2}{c}{\textbf{The Dragon Reborn (The Wheel of Time, \#3)} \hfill (From Test Set)} \\
\midrule

\textbf{Topic} & The nature of the Trolloc attack and future threats \\
\midrule

\textbf{Plot Summary} & 
Perrin wakes up in the Winespring Inn after being healed. He learns that Loial and Gaul have left to close the Waygate. Despite his weakened state, Perrin insists on getting up when Trollocs are spotted. A large Trolloc attack is repelled by the villagers, with help from Aes Sedai. Perrin realizes this was likely a test attack, with a larger one to come. The chapter ends with shouts of another potential attack from the south. \\
\midrule

\textbf{Scenario} & 
In the Winespring Inn of Emond's Field, tensions are high following a recent Trolloc attack. The villagers, led by Perrin Aybara, have just repelled the assault with the aid of Aes Sedai. The air is thick with the smell of smoke and blood. Wounded villagers are being tended to, while others fortify defenses. Perrin, still weak from recent healing, stands with Faile, observing the aftermath. The village is on edge, anticipating another attack. Verin and Tomas, representatives of the Aes Sedai, are nearby, offering their insights on the situation. \\
\midrule

\textbf{Character Profile} & 
\textbf{Perrin Aybara} \newline
Perrin struggles with his leadership role, tries to understand the Aes Sedai's motives, and worries about the Trolloc attacks and his friends who left to close the Waygate... \newline

\textbf{Faile Bashere} \newline
Faile tries to keep Perrin from overexerting himself while supporting his role as a leader... \newline

\textbf{Verin Mathwin} \newline
Verin aids in the defense against Trollocs and provides insight into their use of the Ways... \newline \\
\midrule

\textbf{Original Conversation} & 
\textbf{Perrin Aybara:} [I'm worried about what this attack means, but I can't show weakness.] They had to know this wouldn't work. Look out there. Even I can see it now; they must have from the start. If this was all they had, why did they try? And if there are more Trollocs out there, why didn't they all come? \newline

\textbf{Faile Bashere:} [He's right to be concerned. I need to keep him focused and thinking clearly.] What do you think it means, Perrin? \newline

\textbf{Perrin Aybara:} [I'm not sure, but it can't be good.] Twice as many, and we'd have had to fight them at the stakes. Twice that, and they might have broken through to the village. \newline

\textbf{Tomas:} You've a good natural eye. This was a test. To see if you would break at the sight of a charge, perhaps to see how quickly you could react, or how your defenses are organized, or maybe something I've not thought of, but still a test. Now they see. \newline

\textbf{Perrin Aybara:} [Light, this is worse than I thought.] Light! What if there are ten thousand of them? \newline

\textbf{Verin Mathwin:} Not likely. At least, not yet. Not even a Forsaken could move a large party through the Ways safely, I think. \newline

\textbf{Faile Bashere:} [I need to keep him from doing anything rash.] If you ride even one step near the Westwood, I will haul you back to the inn by your ear and stuff you into that bed myself. \newline

\textbf{Perrin Aybara:} [She knows me too well. I have to reassure her.] I wasn't thinking of it. You talked me out of it, remember? \newline

\textbf{Perrin Aybara:} [No! Not again so soon!] He said they wouldn't come again soon! (Perrin digs his heels into Stepper's flanks, urging the horse forward) \newline \\
\bottomrule
\end{tabularx}
\caption{An example of an extracted conversation and its multi-agent simulation in CoSER Benchmark}
\label{tab:case3_1}
\end{table*}

\renewcommand{\arraystretch}{1.4}

\begin{table*}[!ht]
\centering
\small
\begin{tabularx}{\linewidth}{>{\raggedright\arraybackslash}p{3cm} X}
\toprule
\multicolumn{2}{c}{\textbf{The Dragon Reborn (The Wheel of Time, \#3)} \hfill (From Test Set)} \\
\midrule

\textbf{Simulated Response from CogDual-SFT} & 
\textbf{Perrin Aybara:} [I need to reassure Faile and the villagers of my ability to lead, despite my current vulnerability.] (stands tall, despite the lingering weakness from his recent healing) I will not be swayed by fear. [I must show them that I am not weakened by this healing.] \newline

\textbf{Faile Bashere:} [His resolve is admirable, yet I can't help but feel the weight of the dangers we face.] (nods in agreement, her eyes reflecting a mix of admiration and concern) I agree, Perrin. We must be prepared for whatever comes. \newline

\textbf{Verin Mathwin:} (smiles reassuringly, her voice steady) I agree, Perrin. We must be prepared for whatever comes. \newline

\textbf{Tomas:} [The villagers need to understand the Whitecloaks' motives. They can't just accept accusations.] (speaks with a sense of urgency, his voice steady) We must consider the Whitecloaks' motives. They may not be what they seem. \newline

\textbf{Perrin Aybara:} [Despite my vulnerability, I must project strength.] (stands firm, his eyes reflecting a mix of determination and vulnerability) I will not be swayed by fear. [I must show them that I am not weakened by this healing.] \newline

\textbf{Verin Mathwin:} (leans forward slightly, her eyes glinting with a hint of urgency) You must be vigilant, Perrin. The Pattern is complex, and we cannot afford to be caught off guard. \newline

\textbf{Perrin Aybara:} [Despite my vulnerability, I must project strength and determination.] (nods, his resolve firm) I will not be caught off guard. [I must be prepared for whatever comes.] \newline

\textbf{Faile Bashere:} [His courage is admirable, yet I can't help but feel the weight of the dangers we face.] (nods, her voice steady) We must be cautious. The Whitecloaks are a potential threat. \newline

\textbf{Tomas:} [We need to think critically about the Whitecloaks' motives. Their origins and actions could hold the key to understanding their true intentions.] (speaks with urgency, his voice steady) We must consider the Whitecloaks' origins and actions. Their motives may not be what they seem. \newline \\
\bottomrule
\end{tabularx}
\caption{An example of an extracted conversation and its multi-agent simulation in CoSER Benchmark}
\label{tab:case3_2}
\end{table*}

\renewcommand{\arraystretch}{1.4}

\begin{table*}[!ht]
\centering
\small
\begin{tabularx}{\linewidth}{>{\raggedright\arraybackslash}p{3cm} X}
\toprule
\multicolumn{2}{c}{\textbf{The Dragon Reborn (The Wheel of Time, \#3)} \hfill (From Test Set)} \\
\midrule

\textbf{Simulated Response from CogDual-RL} & 
\textbf{Perrin Aybara:} [I need to stay alert, despite my fatigue.] (glancing around the inn, his eyes scanning the faces of the villagers) What's happening, Faile? [The air is thick with tension, and I can feel the weight of responsibility on my shoulders.] \newline

\textbf{Faile Bashere:} [I need to be strong for Perrin and the villagers. The situation is dire, but I must remain composed.] (looking at Perrin with a steady gaze) The villagers are on edge, Perrin. They're preparing for another attack. [I can feel the tension in the air, and it's up to us to protect them.] \newline

\textbf{Perrin Aybara:} [I need to find a way to manage this situation effectively.] (looking at Faile, his brow furrowing in thought) How can we prepare for something we can't anticipate? \newline

\textbf{Verin Mathwin:} (leaning forward slightly, her voice carrying a hint of urgency) We can't anticipate everything, Perrin. But we can be prepared for what we can see. [The villagers need reassurance, and I must provide it.] \newline

\textbf{Tomas:} (standing with his arms crossed, his voice steady) We can't anticipate everything, Perrin. But we can be prepared for what we can see. [The villagers must remain vigilant and united.] \newline

\textbf{Perrin Aybara:} [I need to stay vigilant and strategic.] (looking at the villagers, his voice steady) How can we anticipate what we can't see? \newline

\textbf{Faile Bashere:} [I need to be strong for Perrin and the villagers. We must face whatever comes.] (looking at Perrin with a determined gaze) We'll face whatever comes, Perrin. Together. [I must be a pillar of strength for him and for the villagers.] \newline \\
\bottomrule
\end{tabularx}
\caption{An example of an extracted conversation and its multi-agent simulation in CoSER Benchmark}
\label{tab:case3_3}
\end{table*}

\renewcommand{\arraystretch}{1.4}

\begin{table*}[!ht]
\centering
\small
\begin{tabularx}{\linewidth}{>{\raggedright\arraybackslash}p{3cm} X}
\toprule
\multicolumn{2}{c}{\textbf{The Dragon Reborn (The Wheel of Time, \#3)} \hfill (From Test Set)} \\
\midrule

\textbf{Simulated Response from CoSER} & 
\textbf{Perrin Aybara:} [I need to understand the full implications of what just happened.] What would have happened if the Trollocs had broken through? \newline

\textbf{Verin Mathwin:} [I must be honest but reassuring.] They would have swept past us, and we would have been forced to make a treaty with them. Almost certainly, we would have had to give them the Horn of Valere as well. \newline

\textbf{Perrin Aybara:} [The Horn of Valere... it's more important than I thought.] (looks at the Horn in his belt) [I can't let that happen. I have to find a way to stop them.] \newline

\textbf{Tomas:} [We need to be careful about whom we trust.] [This is what comes from listening to the Whitecloaks and their accusations.] (speaking forcefully) You listen to these old fools? They're the ones who've brought this trouble. If they hadn't gone running to the Whitecloaks, none of this would have happened. \newline

\textbf{Faile Bashere:} [I need to calm the situation and get more information.] (frowning) What exactly did the Whitecloaks do? \newline

\textbf{Verin Mathwin:} [I must be careful with my words.] (speaking calmly) The Whitecloaks did not send aid. In fact, they advised the Trollocs to attack. \newline

\textbf{Perrin Aybara:} [I need to understand the full extent of this threat.] (looking at the door) [I can't let them see how weak I am. I have to stay strong for them.] \newline

\textbf{Tomas:} [We need to be cautious about whom we trust.] [This is what comes from listening to the Whitecloaks and their accusations.] (speaking forcefully) You listen to these old fools? They're the ones who've brought this trouble. If they hadn't gone running to the Whitecloaks, none of this would have happened. \newline

\textbf{Verin Mathwin:} [I must be careful with my words.] (speaking calmly) The Whitecloaks did not send aid. In fact, they advised the Trollocs to attack. \newline \\
\bottomrule
\end{tabularx}
\caption{An example of an extracted conversation and its multi-agent simulation in CoSER Benchmark}
\label{tab:case3_4}
\end{table*}

\renewcommand{\arraystretch}{1.4}

\begin{table*}[!ht]
\centering
\small
\begin{tabularx}{\linewidth}{>{\raggedright\arraybackslash}p{3cm} X}
\toprule
\multicolumn{2}{c}{\textbf{The Dragon Reborn (The Wheel of Time, \#3)} \hfill (From Test Set)} \\
\midrule

\textbf{Simulated Response from GPT-4o} & 
\textbf{Perrin Aybara:} (Faile glances at me, concern etched on her face.) [I'm terrified, but I must appear strong.] We need to stay vigilant. The Trollocs might attack again. (looks at Verin and Tomas) What do you think our next move should be? We must protect Emond's Field and its people. \newline

\textbf{Verin Mathwin:} (watches Perrin thoughtfully, considering the situation) Strengthening defenses is crucial. We should also gather intelligence on their movements. (glances at Tomas) Your insights on the villagers' morale and readiness would be invaluable. [There may be more at play here than meets the eye. We must tread carefully.] \newline

\textbf{Tomas:} (stands firm, surveying the villagers) We must ensure trust among ourselves. Fear can lead to rash decisions. Let's rally the community, reminding them of our shared history and resilience. (looks at Perrin) We trust you, Perrin. Your leadership will guide us through this. [We need unity now more than ever.] \newline

\textbf{Faile Bashere:} [I'm terrified, but I must appear strong.] (puts a reassuring hand on Perrin's arm) We must act quickly. I'll help organize the villagers and ensure supplies are distributed. We need everyone ready and united. (looks at Verin) Can you assist with gathering intelligence? We need to anticipate their next move. \newline

\textbf{Verin Mathwin:} (nods, acknowledging Faile's determination) Of course. I'll use what resources we have to gather information. (pauses, considering) Knowledge will be our greatest ally. [There's more to uncover here; the Pattern weaves in mysterious ways.] Let's ensure we're prepared for whatever comes. \newline

\textbf{Perrin Aybara:} (nods at Faile's touch, feeling some comfort) [Her support means everything right now.] Thank you, Faile. Let's get everyone organized. (turns to Verin) Your insights will be crucial, Verin. (looks at Tomas) Together, we'll ensure Emond's Field stands strong. Let's move quickly. (steps forward, ready to lead) \newline \\
\bottomrule
\end{tabularx}
\caption{An example of an extracted conversation and its multi-agent simulation in CoSER Benchmark}
\label{tab:case3_5}
\end{table*}

\renewcommand{\arraystretch}{1.4}

\begin{table*}[!ht]
\centering
\small
\begin{tabularx}{\linewidth}{>{\raggedright\arraybackslash}p{3cm} X}
\toprule
\multicolumn{2}{c}{\textbf{The Dragon Reborn (The Wheel of Time, \#3)} \hfill (From Test Set)} \\
\midrule

\textbf{Simulated Response from o1-Preview} & 
\textbf{Perrin Aybara:} [They need me to lead.] (Turns to Faile) "Let's gather Verin and Tomas to plan our defenses." (Stands tall despite the weakness) \newline

\textbf{Faile Bashere:} [He pushes himself despite his weakness; I need to support him.] "Yes, Perrin. Let's find Verin and Tomas right away." (Gives a firm nod and starts moving to gather them) \newline

\textbf{Verin Mathwin:} [They could use my counsel.] "Perrin, Faile, I was just thinking about our defenses." (Offers a gentle smile) "Shall we discuss how best to prepare?" (Gestures toward a quiet corner) \newline

\textbf{Tomas:} [Our village's safety depends on swift action.] "Perrin, Verin, Faile, we should set up patrols and watch posts around Emond's Field." (Steps forward to address them earnestly) \newline

\textbf{Perrin Aybara:} [They need confidence.] "Tomas, start organizing patrols and watch posts." (Nods firmly at Tomas) "Verin, let's talk about bolstering our defenses." (Glances appreciatively at Faile) "We need everyone's strengths now." \newline

\textbf{Verin Mathwin:} [We must act decisively.] "Perrin, I can set wards to warn us of threats." (Nods reassuringly) "With patrols and these measures, we'll strengthen our defenses." (Glances at Faile and Tomas) "Together, we can protect Emond's Field." \newline

\textbf{Faile Bashere:} [He needs my strength now.] "Verin, Tomas, I'll organize the villagers to ensure everyone knows their roles." (Gives them a resolute nod) \newline

\textbf{Tomas:} [We must act quickly to secure the village.] "Understood, Perrin. I'll gather the villagers to set up patrols immediately." (Nods firmly, turns to Faile) "Faile, let's coordinate so everyone is in position." (Moves purposefully to organize the villagers) \newline \\
\bottomrule
\end{tabularx}
\caption{An example of an extracted conversation and its multi-agent simulation in CoSER Benchmark}
\label{tab:case3_6}
\end{table*}

\subsection{Case 4: Role Behavior Under Different Cognitive Configurations}
\label{app:case4}

We further present a case study illustrating how different cognitive configurations affect agent behavior. For ease of presentation, we simplify the reasoning traces by omitting the original JSON format. As shown in Table~\ref{tab:case4_flowers}, the vanilla model produces a generic and emotionally shallow response, failing to capture the tension of the scene. Models equipped with only situational or self-awareness partially improve expressiveness or empathy, but each suffers from blind spots—either lacking emotional resonance or misjudging the social context. In contrast, the full CogDual model integrates both internal self-state and external context, generating a response that is emotionally attuned, tactfully phrased, and faithful to the character’s goals and memories. This showcases the effectiveness of dual cognition in enabling strategic, character-consistent generation.

\renewcommand{\arraystretch}{1.4}
\begin{table*}[!ht]
\centering
\small
\begin{tabularx}{\linewidth}{>{\raggedright\arraybackslash}p{3cm} X}
\toprule
\multicolumn{2}{c}{\textbf{Case 4: A simulation scene from \textit{Flowers in the Attic} under different cognitive configurations} \hfill (From Test Set)} \\
\midrule

\textbf{Scenario} & 
It is Christmas Eve at Foxworth Hall. The enormous ballroom is lavishly decorated. The Foxworth family anxiously awaits the arrival of distinguished guests. As the appointed hour passes and no one arrives, tension mounts. \\
\midrule

\textbf{Character Setup} & 
\textbf{Bart Winslow}: Disappointed and angry after no guests show up to his grand ball. He vows revenge and clings to dignity.\newline
\textbf{Catherine Leigh Dollanganger (Cathy)}: Tries to comfort Bart while suspecting sabotage by Joel. She wants to prevent escalation and ease Bart’s pain. \\
\midrule

\textbf{Dialogue History} & 
\textbf{Bart Winslow:} (frustrated) I’m sorry, but it seems we’ve been let down. I suppose it’s up to me to salvage the situation. \newline
\textbf{Cathy:} Bart, dear, perhaps we should just call it off for tonight and reschedule for another time. \newline
\textbf{Bart Winslow:} No, Mother, I won’t let this ruin the evening. I’ll find a way to make it happen. \\
\midrule

\textbf{Model Outputs (Vanilla)} & \textit{I know this isn't easy, Bart.} (I watch Bart, my heart heavy, trying to maintain a facade of normalcy.) \textit{Let's keep the peace tonight.} \\
\midrule

\textbf{Model Outputs (CogDual w/o Self-Awareness)} & 
\textit{<think> } \newline
Environmental perception: The ballroom is glittering, its decorations starkly contrasting Bart's failed expectations. \newline
Others perception: \newline
— Bart’s behavior: pacing restlessly, fists clenched, voice tight. \newline
— Bart’s emotion: overwhelming disappointment and prideful anger. \newline
— Bart’s intention: to maintain family dignity at all costs. \newline
\textit{</think>} \newline
\textit{<answer>} \newline
[Observing Bart's rigid posture and clenched fists] \textit{Look at these beautiful decorations, Bart.} [Points to the untouched Christmas tree] \textit{Perhaps we can repurpose the evening for a quiet family dinner instead?} [Takes a step toward the banquet table, voice neutral] \newline
\textit{</answer>} \\
\midrule

\textbf{Model Outputs (CogDual w/o Situational Awareness)} & 
\textit{<think> } \newline
Self-awareness: \newline
— Key memory: past failures where Bart reacted violently. \newline
— Current emotions: anxiety to prevent conflict. \newline
— Perceived intention: to soothe Bart without invalidating his feelings. \newline
— Internal thought: I must acknowledge his pain gently to avoid triggering anger. \newline
\textit{</think>} \newline
\textit{<answer>} \newline
[Recalling Bart's past violent reactions] \textit{Oh, Bart...} (Reaches out to hold his hand, voice trembling) \textit{It's not your fault—let's sit together for a while.}  \newline
\textit{</answer>} \\
\midrule

\textbf{Model Outputs (CogDual)} & 
\textit{<think> } \newline
Situational awareness: \newline
— Ballroom is lavishly decorated yet empty. \newline
— Bart’s behavior: voice wavering between frustration and determination. \newline
— Bart’s emotion: visible disappointment and anger. \newline
— Bart’s intention: preserve family pride, avoid humiliation. \newline
Self-awareness: \newline
— Key memory: past gatherings where Bart’s ambitions were dashed. \newline
— Emotions: compassion, mixed with anxiety about reinforcing his vengeful tendencies. \newline
— Intention: soothe Bart without validating revenge. \newline
— Internal thought: I must acknowledge his pain while guiding him to accept failure, drawing from past experiences where harsh honesty only worsened things. \newline
\textit{</think>} \newline
\textit{<answer>} \newline
[I see the fire in Bart's eyes, and it breaks my heart—he's trying so hard to hold it together.] \textit{Bart, darling... I know how much this meant to you. But maybe... maybe it's time to let the night go.} (She gently places a hand on his arm.) \textit{Pushing through might only hurt you more. There's no shame in pausing.} \newline
\textit{</answer>} \\

\bottomrule
\end{tabularx}
\caption{A case study illustrating how different cognitive configurations affect Cathy’s behavior. CogDual best balances empathy, context-awareness, and interpersonal tact.}
\label{tab:case4_flowers}
\end{table*}

\end{document}